\documentclass[12pt,a4paper]{article}

\usepackage[T1]{fontenc}
\usepackage[utf8]{inputenc}
\usepackage{amsmath,amssymb}
\usepackage{newtxtext,newtxmath}

\usepackage[margin=1in]{geometry}
\usepackage{graphicx}
\usepackage{pdfpages}
\usepackage{booktabs}
\usepackage{multirow}
\usepackage{array}
\usepackage{threeparttable}
\usepackage{tabularx}
\usepackage{makecell}
\usepackage{float}
\usepackage{xcolor}
\usepackage[numbers,sort&compress]{natbib}
\usepackage[hidelinks]{hyperref}
\usepackage{parskip}
\title{Learning Human Health and Diseases from 24-hour Wrist Movement}
\date{}
\author{%
\centering
\linespread{1.05}\selectfont
Yong~Wang$^{1,2\ast\dagger}$,
Dylan~McGagh$^{3,4,5}$,
Katya~Broomberg$^{3,6}$,
Zizheng~Zhang$^{3,5,6}$,
Jonathan~Carter$^{7}$,
Junayed~Naushad$^{8}$,
Laura~Brocklebank$^{3,6}$,
Yang~Sun$^{3,9}$,
George~Nicholson$^{3,9}$,
Dianjianyi~Sun$^{10,11,12}$,
Canqing~Yu$^{10,11,12}$,
Jun~Lv$^{10,11,12}$,
Maxim~Barnard$^{6}$,
Hubert~Lam$^{6}$,
Andrew~Steptoe$^{13}$,
David~W.~Eyre$^{3,5,6}$,
Liming~Li$^{10,11,12}$,
Zhengming~Chen$^{6}$,
Naomi~Wray$^{1,2}$,
Spiros~Denaxas$^{14,15,16,17}$,
Gary~S.~Collins$^{18,19}$,
Huaidong~Du$^{6}$,
Aiden~Doherty$^{2,3,6\ast\ddagger}$,
Hang~Yuan$^{3,6\ast\dagger\ddagger}$\\[1.5em]
\begingroup
\small
\linespread{1}\selectfont
\setlength{\parskip}{0pt}
\begin{flushleft}
$^{1}$Department of Psychiatry, University of Oxford, Oxford, UK.\\
$^{2}$Pioneer Centre for SMARTbiomed, Oxford, UK.\\
$^{3}$Big Data Institute, University of Oxford, Oxford, UK.\\
$^{4}$Nuffield Department of Orthopaedics, Rheumatology and Musculoskeletal Sciences, University of Oxford, Oxford, UK.\\
$^{5}$Oxford University Hospitals, Oxford, UK.\\
$^{6}$Nuffield Department of Population Health, University of Oxford, Oxford, UK.\\
$^{7}$Department of Engineering Science, University of Oxford, Oxford, UK.\\
$^{8}$Department of Computer Science, University of Oxford, Oxford, UK.\\
$^{9}$Department of Statistics, University of Oxford, Oxford, UK.\\
$^{10}$Department of Epidemiology and Biostatistics, School of Public Health, Peking University Health Science Center, Beijing, China.\\
$^{11}$Peking University Center for Public Health and Epidemic Preparedness and Response, Beijing, China.\\
$^{12}$Key Laboratory of Epidemiology of Major Diseases (Peking University), Ministry of Education, Beijing, China.\\
$^{13}$Department of Behavioural Science and Health, University College London, London, UK.\\
$^{14}$Institute of Health Informatics, University College London, London, UK.\\
$^{15}$Interdisciplinary Transformation University (ITU), Linz, Austria.\\
$^{16}$NIHR Biomedical Research Centre at University College London Hospitals (UCLH) NHS Foundation Trust, London, UK.\\
$^{17}$National and Kapodistrian University of Athens, Athens, Greece.\\
$^{18}$Department of Applied Health Sciences, School of Health Sciences, College of Medicine and Health, University of Birmingham, Birmingham, UK.\\
$^{19}$National Institute for Health and Care Research (NIHR) Biomedical Research Centre: Birmingham, University Hospitals Birmingham NHS Foundation Trust and University of Birmingham, Birmingham, UK.\\[0.7em]
$^{\ast}$Corresponding authors. Emails: yong.wang@psych.ox.ac.uk; aiden.doherty@ndph.ox.ac.uk; hang.yuan@ndph.ox.ac.uk.\\
$^{\dagger}$These authors contributed equally to this work.\\
$^{\ddagger}$These authors jointly supervised this work.
\end{flushleft}
\endgroup
}

\makeatletter
\renewcommand{\@maketitle}{%
  \newpage
  \null
  \vskip 2em%
  \begin{center}%
    {\LARGE\bfseries \@title\par}%
    \vskip 1.5em%
    {\large \@author\par}%
  \end{center}%
  \par
  \vskip 1.5em%
}
\makeatother

\begin{document}
\maketitle
\clearpage

\begin{abstract}
Much of human health and function unfolds beyond the clinic, through the movements of everyday life. Wrist-worn accelerometers capture these movements continuously, yet their rich signals are often reduced to a small set of predefined behavioural summary measures. Here, we present Sensori, a self-supervised foundation model that learns general-purpose health representations directly from 24 hours of raw tri-axial wrist movement. We developed and evaluated the model across four population-based cohorts from the United Kingdom, China and the United States, comprising 122,640 participants contributing 683,617 person-days of free-living recordings. Sensori condensed each day of movement into a representation that captured diverse movement behaviours, demographic characteristics, health axes and physical function. Evaluation in independent cohorts showed that these representations generalised across populations and measurement settings without retraining. When added to common clinical covariates, Sensori significantly improved prevalent disease classification for 52 of 102 eligible conditions (median delta AUROC, 0.060; range, 0.012--0.242) and incident disease risk prediction for 26 of 87 eligible conditions (median delta Uno's C-index, 0.064; range, 0.025--0.172), with the largest gains for neurological and psychiatric disorders. These findings establish 24-hour wrist movement as a rich and scalable source of health information, with the potential to support passive health monitoring and disease prediction at population scale.
\end{abstract}

\newpage
\section*{Introduction}\label{sec1}
Human bipedalism freed the upper limbs from locomotion~\cite{darwin1871descent}. 
This evolutionary transition transformed the hands into the primary interface through which humans interact with their environment. 
Nearly every aspect of daily life, including eating, working and caregiving, is expressed through coordinated hand and wrist movements. 
Consequently, a 24-hour recording of wrist movement provides a continuous window into how an individual lives and functions in the real world.

The widespread adoption of wrist-worn accelerometers has made continuous measurement of 24-hour movement increasingly accessible. 
As of 2026, approximately one in three online adults worldwide owns a smartwatch capable of continuously capturing wrist movement~\cite{kemp2025digital}.
Wrist-worn accelerometers have also been deployed in large-scale population-based cohorts, enabling the study of associations between free-living movement behaviours and diverse health outcomes~\cite{doherty2017large, chen2023device, cdc_nhanes_pax80g_2026, cdc_nhanes_pax80h_2026, master2022association}.

Human behaviour is continuous, hierarchical and multidimensional, spanning temporal scales from second-level movement dynamics to daily behavioural routines. 
Device-measured behaviour has substantially refined our understanding of the associations between physical activity and health, revealing stronger links with longevity than those estimated from self-reported measures~\cite{wasfy2022examining}.
Recent studies have further linked wrist-derived movement behaviours with mortality and a broad range of chronic diseases, including cancer, cardiovascular and neurological disorders~\cite{stamatakis2022association, schalkamp2023wearable, master2022association, shreves2025amount}. 
However, these studies have largely relied on a limited set of already known behavioural summary measures, such as activity intensity, step counts and sleep duration, rather than modelling 24-hour wrist movement as a continuous behavioural time series. 
As a result, much of the temporal organisation and behavioural context embedded in daily movement remains unexplored.

Meanwhile, healthcare systems are aiming to transition towards personalised, preventive and home-based care, creating growing demand for objective measures that characterise health continuously in everyday life. 
Passive wearable sensing is expected to play a central role in this transition, with wrist-worn accelerometers particularly well suited owing to their low cost, ease of use, widespread adoption and ability to capture free-living behaviour at scale. 
This direction is increasingly reflected in national healthcare strategies, including the United Kingdom’s commitment to integrate wearable technologies into routine care by 2035~\cite{dhsc2025fit, lu2026wearables}. 
Realising this vision, however, requires methods that can transform continuous behavioural measurements into clinically meaningful representations of health and disease.

Self-supervised learning has underpinned recent advances in modelling complex biosignals, including glucose~\cite{lutsker2026foundation}, cardiac activity~\cite{abbaspourazad2023large, pillai2025papagei}, sleep~\cite{thapa2026multimodal} and gait~\cite{gabet2026gait}, proving its ability to uncover clinically informative features from large amounts of unlabelled data. 
Here, we introduce Sensori, a movement foundation model trained through self-supervision on 24-hour wrist movement.
By modelling a complete 24-hour recording of raw tri-axial wrist acceleration directly, Sensori captures temporal information that is not represented by coarse predefined behavioural summaries. 
We show that its learned representations characterise health state and physical function across diverse assessment tasks, generalise to independent cohorts without retraining, and improve disease prediction for conditions with movement-related phenotypes.
These findings establish continuous wrist movement as a rich, general-purpose signal for measuring health and disease in everyday life.

\subsection*{Learning health representations from 24-hour wrist movement}

\begin{figure}[H]
\centering
\includegraphics[width=\textwidth]{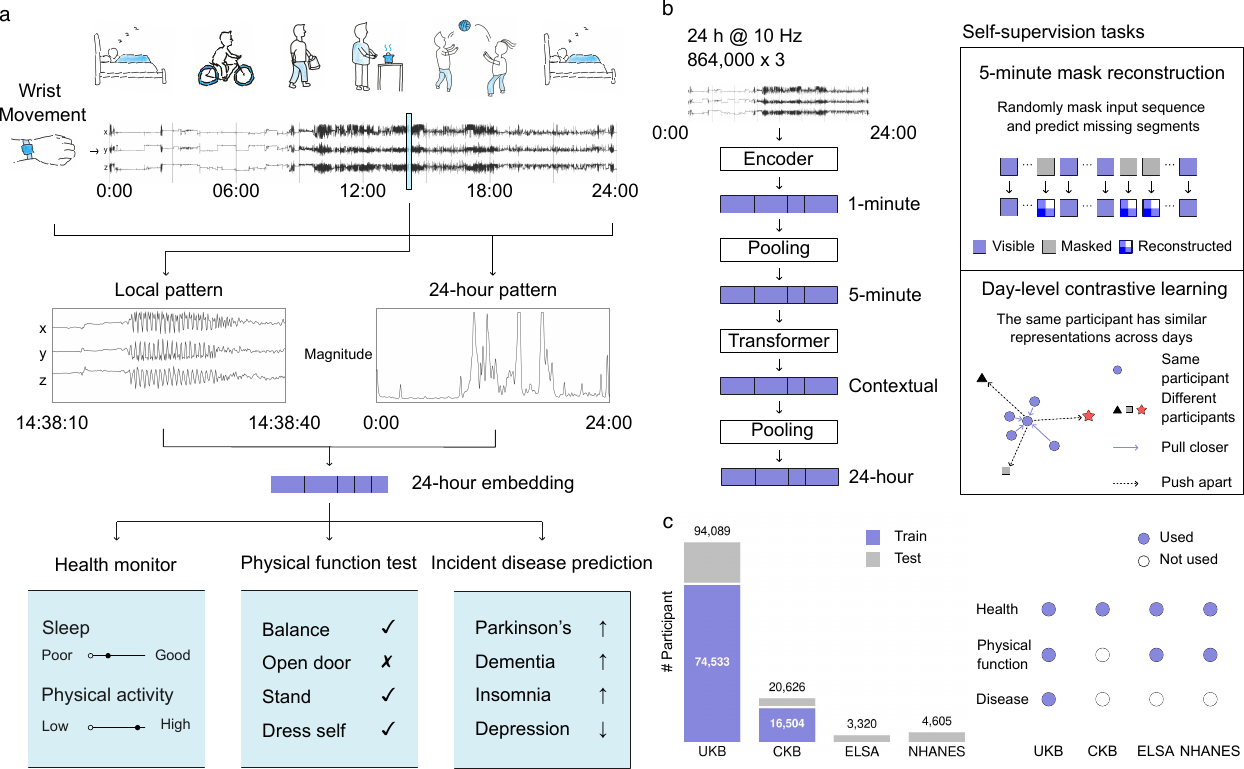}
\caption{\textbf{Study overview.}
\textbf{a}, Conceptual framework of Sensori. 
Sensori captures patterns in raw tri-axial wrist movement across multiple temporal scales over a 24-hour period and integrates them into a day-level health representation for downstream tasks.
\textbf{b}, Sensori architecture and self-supervised pretraining. 
Sensori takes 24 hours of raw 10-Hz wrist movement as input and generates a day-level embedding using a multiscale deep neural network. 
Sensori is pretrained using masked reconstruction at the five-minute level and contrastive learning at the day level.
\textbf{c}, Study cohorts and evaluation design. 
Left, participant partitioning for pretraining and downstream evaluation. Right, cohorts used for each downstream evaluation task.}
\label{fig:fig1}
\end{figure}

A full day of tri-axial wrist movement contains information about an individual’s health and disease status across multiple temporal scales. 
Local movement patterns, such as wrist tremor and gait asymmetry, can indicate neurological disorders, whereas longer-term patterns, such as prolonged physical inactivity and episodes of restless sleep, can reflect underlying health conditions. 
Sensori is a deep neural network that integrates these multiscale signals from a full-day recording into a day-level health representation. This representation supports diverse downstream tasks, including health-state monitoring, physical function assessment and disease prediction (Fig.~\ref{fig:fig1}a).

At a sampling frequency of 10 Hz, a 24-hour recording contains 864,000 observations per axis, yielding a sequence substantially longer than the typical inputs used by existing movement models. 
To process sequences of this length, Sensori uses a multiscale architecture in which pooling operations progressively reduce the temporal resolution. 
A convolutional encoder first produces one embedding vector per minute. 
Mean pooling then generates five-minute embeddings, which are passed to a transformer to model temporal dependencies across the day. 
A second mean-pooling operation over the transformer outputs produces a 768-dimensional day-level representation (Fig.~\ref{fig:fig1}b, left).

Sensori is pretrained using two complementary objectives designed to capture movement patterns at different temporal scales: masked reconstruction and day-level contrastive learning. 
We selected this combination after systematically evaluating alternative pretraining strategies (Supplementary Fig.~\ref{fig:scale_task} and Supplementary Tables~\ref{tab:supp_model}--\ref{tab:epoch99_model_performance_extended}). For masked reconstruction, a random subset of the five-minute embeddings provided to the transformer is masked, and the model reconstructs these embeddings from the remaining inputs, thereby learning contextual relationships among movement segments within a day. 
Day-level contrastive learning encourages representations from different days recorded from the same individual to be more similar to each other than to representations from different individuals, promoting the capture of movement characteristics that are stable across days (Fig.~\ref{fig:fig1}b, right).

Our study draws on four population-based cohorts from the United Kingdom, China and the United States: the UK Biobank (UKB)~\cite{doherty2017large}, the China Kadoorie Biobank (CKB)~\cite{chen2023device}, the English Longitudinal Study of Ageing (ELSA)~\cite{glag197} and the National Health and Nutrition Examination Survey (NHANES)~\cite{cdc_nhanes_pax80g_2026, cdc_nhanes_pax80h_2026}. 
Each participant wore a wrist-worn accelerometer for up to seven days in UKB, CKB and NHANES, and up to eight days in ELSA. 
After quality control, 122,640 participants contributed 683,617 person-days of wrist accelerometer recordings: 94,089 from UKB, 20,626 from CKB, 3,320 from ELSA and 4,605 from NHANES (Supplementary Figs.~\ref{fig:supp_ukb_participant_inclusion}--\ref{fig:supp_nhanes_participant_inclusion}). Participants in all cohorts except NHANES had a median age above 60 years; females outnumbered males, and median body mass index (BMI) was at least 24.3 $kg/m^2$. 
Device-measured movement behaviours also differed across cohorts. 
CKB participants were the most active (median daily step count, \(10{,}105\); 25th--75th percentiles, \(7{,}306\)--\(13{,}165\)) and had the shortest sleep duration (median, \(6.03\) hours; 25th--75th percentiles, \(5.23\)--\(6.80\) hours), whereas UKB participants were the second most active (median daily step count, \(7{,}938\); 25th--75th percentiles, \(6{,}007\)--\(10{,}126\)) and had the longest sleep duration (median, \(6.86\) hours; 25th--75th percentiles, \(6.24\)--\(7.42\) hours).
Cohort-specific study periods, accelerometry protocols and population characteristics are provided in Table~\ref{tab:population_characteristics}.

The pretraining dataset comprised approximately 80\% of participants from UKB and CKB (\(n=91{,}037\)), providing 502,823 person-days of accelerometer recordings. The remaining 20\% of participants from UKB and CKB, together with all eligible participants from ELSA and NHANES (\(n=31{,}603\)), were reserved for downstream evaluation and provided 180,794 person-days of recordings (Fig.~\ref{fig:fig1}c, left). Health-state monitoring was evaluated in all four studies, whereas physical function assessment was evaluated in UKB, ELSA and NHANES (Supplementary Tables~\ref{tab:covariates_ukb}--\ref{tab:covariates_nhanes}). Disease-related tasks were evaluated only in UKB because disease phenotyping required the harmonisation of multiple data sources and incident disease risk prediction required longitudinal follow-up (Fig.~\ref{fig:fig1}c, right).

Throughout this study, we primarily used the mean and standard deviation of acceleration as benchmarks because they are direct measures of movement intensity. For health evaluation and disease-risk prediction, we also included broader sets of device-measured movement behaviours commonly used in wearable applications. Because these measures are subject to greater measurement error, acceleration statistics provide a more readily interpretable benchmark~\cite{doherty2020accelerometer,small2024self,yuan2024self}.

\subsection*{Capturing diverse movement behaviours}
\begin{figure}[H]
\centering
\includegraphics[width=\textwidth]{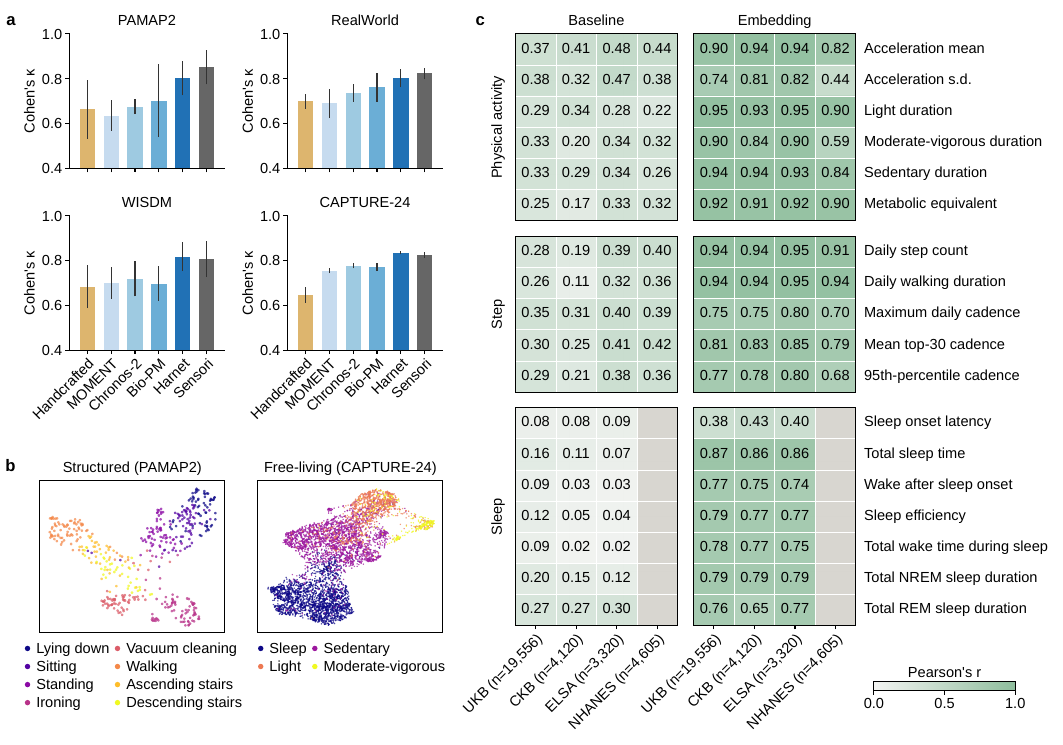}
\caption{\textbf{Sensori representations capture diverse behavioural traits across temporal scales and populations.}
\textbf{a}, Linear-probe performance for one-minute activity recognition across four external benchmark datasets. 
Sensori is compared with handcrafted features, general-purpose time-series foundation models (MOMENT and Chronos-2) and domain-specific models (Bio-PM and Harnet). 
Bars and error bars indicate mean Cohen’s $\kappa$ and s.d. across five participant-wise cross-validation folds, respectively.
\textbf{b}, Uniform manifold approximation and projection (UMAP) plots of minute-level Sensori embeddings under structured (PAMAP2) and free-living (CAPTURE-24) conditions. 
Each point represents a one-minute window coloured by its activity label; CAPTURE-24 is randomly subsampled to 5,000 windows for visualisation.
\textbf{c}, Cross-cohort prediction of day-level device-measured behavioural traits and device statistics. 
Linear probes are fitted on the UKB training set and evaluated without refitting in the UKB test set, CKB test set, ELSA and NHANES. 
The demographic baseline uses age, sex and BMI, whereas the embedding model uses Sensori 24-hour embeddings. 
Cell values show Pearson’s $r$ between predicted and target values; cell colour represents values on a scale from 0 to 1, and grey cells indicate unavailable traits.
\(n\), number of participants.}
\label{fig:2}
\end{figure}

A general representation of 24-hour wrist movement should preserve behaviourally relevant information across temporal scales. 
Accordingly, we first evaluated Sensori’s minute-level embeddings on four external human activity recognition benchmarks: PAMAP2~\cite{reiss2012introducing}, RealWorld~\cite{sztyler2016body}, WISDM~\cite{weiss2019wisdm} and CAPTURE-24~\cite{chan2024capture24} (Supplementary Tables~\ref{tab:har_datasets}--\ref{tab:har_class_distribution}).
Using linear probing, we compared Sensori with five baselines: handcrafted features; two general-purpose time-series foundation models, MOMENT~\cite{goswami2024moment} and Chronos-2~\cite{ansari2025chronos}; and two domain-specific models, Bio-PM~\cite{tarale2026bio} and Harnet~\cite{yuan2024harnet}.
Although Sensori was not specifically designed for human activity recognition, its minute-level embeddings ranked first in mean Cohen’s \(\kappa\) on PAMAP2 (mean \(\kappa\), 0.852; standard deviation (s.d.), 0.077) and RealWorld (mean \(\kappa\), 0.824; s.d., 0.026), and second on WISDM (mean \(\kappa\), 0.807; s.d., 0.081) and CAPTURE-24 (mean \(\kappa\), 0.826; s.d., 0.012) in five-fold participant-wise cross-validation (Fig.~\ref{fig:2}a).
On the latter two datasets, Sensori was surpassed only by Harnet, our previous model developed for human activity recognition.
Corresponding macro F1 scores are reported in Supplementary Fig.~\ref{fig:supp_har}a.
Uniform manifold approximation and projection (UMAP) plots of Sensori’s minute-level embeddings showed separation among activities with distinct movement intensities and patterns under both structured and free-living conditions (Fig.~\ref{fig:2}b and Supplementary Fig.~\ref{fig:supp_har}b). Details of data preprocessing and method implementation for these benchmarks are provided in Supplementary Note~\ref{supp_note_har}.

At the 24-hour scale, we next evaluated whether Sensori embeddings captured diverse movement behavioural traits across populations, spanning six commonly used physical-activity measures, five step measures and seven sleep measures~\cite{doherty2020accelerometer, small2024self, yuan2024self}. Definitions of these traits are provided in Supplementary Table~\ref{supp_tab:device_variable_definitions}.
Linear probes were fitted in the UKB training set and evaluated without refitting across four sets: the held-out UKB and CKB test sets, ELSA and NHANES. 
We compared the performance of Sensori embeddings with that of a demographic baseline comprising age, sex and BMI. 

Across all cohorts, Sensori embeddings outperformed the demographic baseline in predicting device-measured behavioural traits and device statistics (Fig.~\ref{fig:2}c).
Pearson’s \(r\) ranged from 0.02 to 0.48 for the demographic baseline and from 0.38 to 0.95 for Sensori, indicating that the embeddings captured information about movement behaviours beyond that available from age, sex and BMI alone.
Because the idle sleep mode of the NHANES accelerometer suppresses the low-movement information needed to reliably derive sleep-related traits, we did not evaluate sleep traits in this cohort.
The corresponding 95\% confidence intervals (CIs) are provided in Supplementary Table~\ref{supp_tab:device_trait_ci}.

\subsection*{Encoding key demographic characteristics and multidimensional health axes}
Beyond conventional movement behaviours, daily movement patterns may reflect broader inter-individual differences in demographic and health characteristics. 
We therefore tested whether these characteristics could be inferred from Sensori representations across cohorts. 
Linear probes were fitted in the UKB training set and evaluated without refitting in the held-out UKB and CKB test sets and in ELSA and NHANES. 
Sex was evaluated as a binary classification task, whereas age and BMI were evaluated as regression tasks. 
For comparison, we fitted the same linear models using the mean and standard deviation of acceleration.

\begin{figure}[H]
\centering
\includegraphics[width=\textwidth]{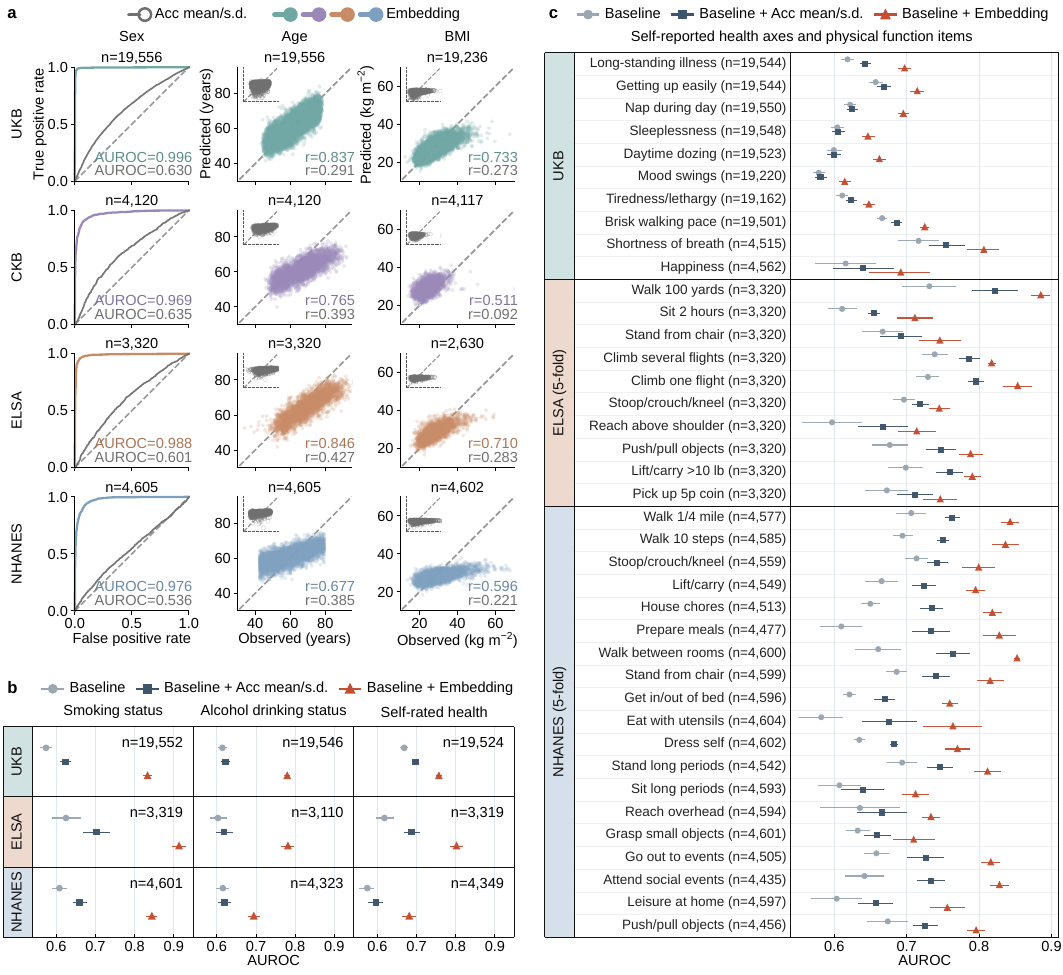}
\caption{\textbf{Sensori encodes key demographic characteristics and multidimensional health axes.}
\textbf{a}, Prediction of sex, age and BMI in held-out evaluation sets from UKB, CKB, ELSA and NHANES, shown as receiver operating characteristic curves with AUROC for sex and scatter plots with Pearson’s $r$ for age and BMI. 
Predictions from day-level Sensori representations are shown in colour, with one colour per cohort, and predictions based on the mean and standard deviation of acceleration (Acc mean/s.d.) are shown in grey.
\textbf{b}, AUROC for three harmonised traits---smoking status, alcohol drinking status and self-rated health---across UKB, ELSA and NHANES. 
These traits were selected for their relevance to health and availability in all three cohorts. 
Points and error bars indicate AUROC and 95\% bootstrap CIs, respectively.
\textbf{c}, AUROC for self-reported health axes in UKB and physical function items in ELSA and NHANES. 
For UKB, linear probes were fitted on the UKB training set and evaluated in the held-out UKB test set; points and error bars indicate AUROC and 95\% bootstrap CIs, respectively.
For ELSA and NHANES, linear probes were evaluated using within-cohort five-fold cross-validation; points and error bars indicate mean AUROC and s.d. across the five outer folds, respectively.
Baseline denotes age, sex and BMI; Acc mean/s.d. denotes the mean and standard deviation of acceleration; Embedding denotes Sensori day-level embeddings.
\(n\), number of participants.}
\label{fig:3}
\end{figure}

Sensori embeddings accurately classified sex and generalised well across cohorts, achieving an area under the receiver operating characteristic curve (AUROC) of 0.996 in UKB, 0.969 in CKB, 0.988 in ELSA and 0.976 in NHANES (Fig.~\ref{fig:3}a).
The embeddings were also predictive of age and, to a lesser extent, BMI.
Pearson’s \(r\) values for age and BMI were 0.837 and 0.733, respectively, in UKB; 0.765 and 0.511 in CKB; 0.846 and 0.710 in ELSA; and 0.677 and 0.596 in NHANES. 
By contrast, models using only the mean and standard deviation of acceleration showed substantially lower performance: AUROC ranged from 0.536 to 0.635 for sex, whereas Pearson’s \(r\) ranged from 0.291 to 0.427 for age and from 0.092 to 0.283 for BMI. 
The corresponding 95\% CIs are provided in Supplementary Table~\ref{supp_tab:age_sex_BMI_ci}.

We next tested whether health-related information encoded by Sensori transferred across cohorts. 
Smoking status, alcohol drinking status and self-rated health were harmonised across UKB, ELSA and NHANES (Fig.~\ref{fig:3}b). These traits were selected for their relevance to health and availability in all three cohorts.
For each trait, linear probes were fitted only in the UKB training set and applied without refitting to the held-out UKB test set, ELSA and NHANES. 
We compared three models: a demographic baseline comprising age, sex and BMI; a second model combining the demographic variables with the mean and standard deviation of acceleration; and a third model combining the demographic variables with the day-level Sensori embeddings. 
Sensori achieved the highest AUROC for all nine cohort–trait combinations. 
Improvement was particularly consistent for smoking status, with AUROCs of 0.833 (95\% CI, 0.821--0.844) in UKB, 0.913 (95\% CI, 0.895--0.932) in ELSA and 0.844 (95\% CI, 0.830--0.858) in NHANES.
These consistent improvements suggest that health-related information captured from 24-hour wrist movement is preserved across cohorts.

For the cohort-specific evaluations, protocols differed according to the available health axes and physical function items. 
Probes for the ten UKB health axes were fitted in the UKB training set and evaluated in the held-out UKB test set. 
The ELSA and NHANES physical function items, which are commonly used to assess limitations in daily activities, were evaluated separately using within-cohort five-fold cross-validation.
Across all items, AUROC estimates were higher for Sensori than for both comparator models (Fig.~\ref{fig:3}c).
The highest mean AUROC values were observed for mobility-related activities, including walking 100 yards in ELSA (mean, 0.885; s.d., 0.013) and walking between rooms in NHANES (mean, 0.852; s.d., 0.005).
Notably, the predictive signal extended beyond mobility to activities such as preparing meals (mean, 0.828; s.d., 0.023) and attending social events (mean, 0.828; s.d., 0.014) in NHANES.
These findings indicate that 24-hour wrist movement captures information about both physical capacity and its expression in everyday function.

\subsection*{Stability of Sensori representations}

\begin{figure}[H]
\centering
\includegraphics[width=\textwidth]{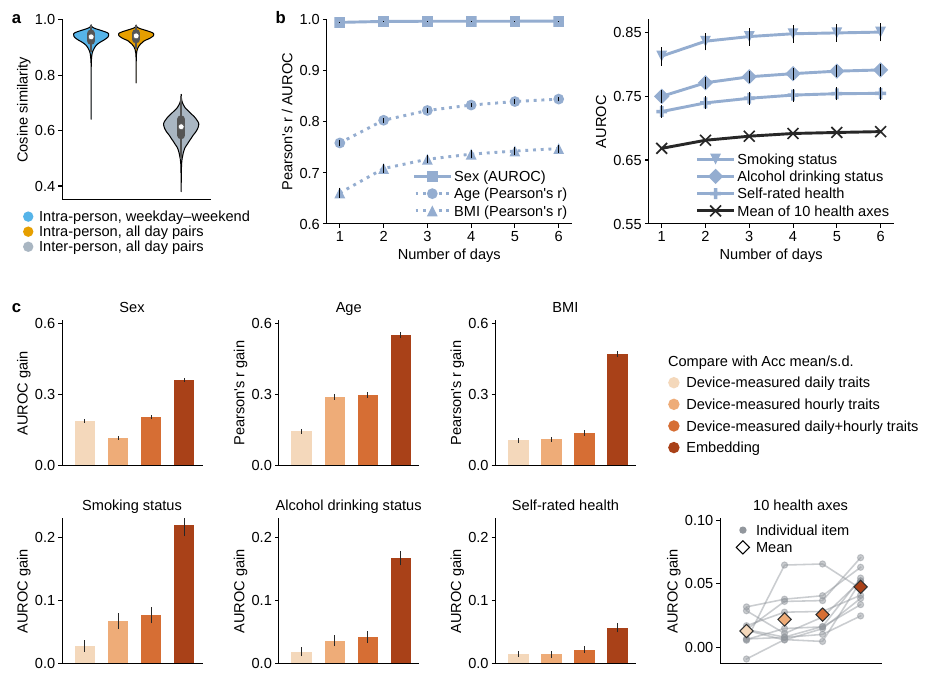}
\caption{\textbf{Stability across days, performance with multi-day inputs and comparison with conventional device-measured behavioural traits.}
\textbf{a}, Participant-level cosine similarity of day-level embeddings within and between individuals. 
Central dots, bars and whiskers represent the median, interquartile range and most extreme values within 1.5\(\times\) the interquartile range, respectively; violin widths indicate density.
\textbf{b}, Linear-probe performance of day-level embeddings using one to six input days at test time. 
Markers and error bars indicate performance and 95\% bootstrap CIs, respectively.
The unweighted mean across ten UKB health axes is shown without a CI.
\textbf{c}, Gain in linear-probe performance relative to corresponding models using the mean and standard deviation of acceleration (Acc mean/s.d.).
In the six task-specific plots, bars show performance gains using device-measured daily traits, hourly traits, their combination and Sensori day-level embeddings.
Error bars indicate 95\% CIs estimated by paired bootstrap resampling of participants.
The rightmost plot shows AUROC gains for each of ten health axes (circles connected by lines) and their unweighted means (diamonds). 
All analyses were conducted using the held-out UKB test set.}
\label{fig:fig4}
\end{figure}

We next examined the stability of Sensori representations across recording days and the additional information gained by aggregating multiple days. 
In the held-out UKB test set, we computed pairwise cosine similarities between day-level embeddings from different days within the same participant, between weekday and weekend days within the same participant, and between days from different participants. 
Day-level embeddings were substantially more similar within than between participants (Fig.~\ref{fig:fig4}a).
Similarities were comparable across all within-participant day pairs (median, 0.941; 25th--75th percentiles, 0.929--0.952) and weekday--weekend pairs (median, 0.938; 25th--75th percentiles, 0.923--0.950), but markedly lower between participants (median, 0.614; 25th--75th percentiles, 0.583--0.641).

Despite this within-participant stability, we found that aggregating additional recording days improved downstream health prediction.
We evaluated linear probes using cumulative averages of each participant’s first one through six valid day-level embeddings. 
Predictive performance improved across all downstream tasks as the number of input days increased (Fig.~\ref{fig:fig4}b).
The health-axis curve represents the unweighted mean across the ten axes; individual trajectories are shown in Supplementary Fig.~\ref{supp_fig:health_axes}a.
Gains were generally largest between one and two input days and diminished as the number of days increased.
One exception was sex classification, for which performance was already near-perfect with a single input day (AUROC \(=0.995\)).
Although the magnitude of improvement varied by task, the overall pattern indicated that multi-day recordings contained predictive information beyond that available from a single 24-hour recording.

Finally, we compared Sensori with a comprehensive set of conventional behavioural traits derived from wrist accelerometry. 
The daily traits spanned physical activity, step and sleep, as introduced in Fig.~\ref{fig:2}c. 
To provide a more stringent comparison, we additionally derived hour-of-day summaries of these traits where available and evaluated the daily and hourly traits separately and in combination. 
Conventional behavioural traits improved downstream performance over a baseline using the mean and standard deviation of acceleration, and incorporating hourly information generally provided further gains; however, the combined daily and hourly traits remained less predictive than Sensori embeddings (Fig.~\ref{fig:fig4}c). 
Across the ten health axes, the mean AUROC improvement relative to the acceleration baseline was 0.026 for the combined daily and hourly traits and 0.048 for Sensori embeddings. 
Results for the individual health axes are provided in Supplementary Fig.~\ref{supp_fig:health_axes}b.

\subsection*{Disease prediction}

Having established that Sensori encodes diverse behavioural traits and health variation, we examined whether these representations also captured signals of prevalent disease and future disease risk in the UKB test set.
Outcomes were defined using a combination of self-reported information, primary care records and hospital records and harmonised using the pomegranate and CALIBER phenotyping frameworks~\cite{denaxas2019uk, torralbo2025computational}. 
Case counts for each condition are reported in Supplementary Table~\ref{tab:supp_disease_inclusion}. 
We included conditions with at least 60 cases in each analysis. 
The number of prevalent cases ranged from 60 to 48,019, with a median of 373, whereas the number of incident cases ranged from 60 to 9,084, with a median of 288. 
For both tasks, we used a clinical covariate model comprising age, sex, BMI, ethnic background, alcohol consumption and smoking status as the baseline. 
Acceleration features or fine-tuned Sensori embeddings (Supplementary Table~\ref{tab:pilot_full_sft_auroc}) were then added to these clinical covariates (Methods).

For prevalent disease classification, adding Sensori embeddings to the clinical covariate model significantly improved AUROC for 52 of 102 eligible conditions across the six disease categories and significantly reduced AUROC for three (Supplementary Table~\ref{tab:fig6_paired_bootstrap_summary}).
The proportion of conditions with significant improvements was highest for neurological (9 of 11) and psychiatric conditions (8 of 10), followed by musculoskeletal (12 of 21), respiratory (7 of 15), cardiovascular (13 of 35) and endocrine conditions (3 of 10).
The largest improvements were observed for multiple sclerosis (delta AUROC, 0.242; 95\% CI, 0.174--0.308), essential tremor (delta AUROC, 0.240; 95\% CI, 0.135--0.345), Parkinson’s disease (delta AUROC, 0.223; 95\% CI, 0.143--0.298), schizophrenia, schizotypal and delusional disorders (delta AUROC, 0.220; 95\% CI, 0.108--0.326) and bipolar affective disorder and mania (delta AUROC, 0.199; 95\% CI, 0.129--0.273).

For 6-year incident disease risk prediction, adding Sensori embeddings significantly improved Uno’s concordance index (C-index) for 26 of 87 eligible conditions, with no significant reductions (Supplementary Table~\ref{tab:fig6_paired_bootstrap_summary}).
Significant improvements were most frequent for psychiatric (7 of 7), neurological (3 of 7), musculoskeletal (6 of 19), respiratory (4 of 15), endocrine (2 of 8) and cardiovascular conditions (4 of 31).
The largest improvements were observed for Parkinson’s disease (delta C-index, 0.172; 95\% CI, 0.102--0.241), essential tremor (delta C-index, 0.171; 95\% CI, 0.077--0.279), spondylolisthesis (delta C-index, 0.116; 95\% CI, 0.058--0.182), depression (delta C-index, 0.114; 95\% CI, 0.047--0.178) and alcohol problems (delta C-index, 0.102; 95\% CI, 0.051--0.154).

\begin{figure}[H]
\centering
\includegraphics[width=\textwidth]{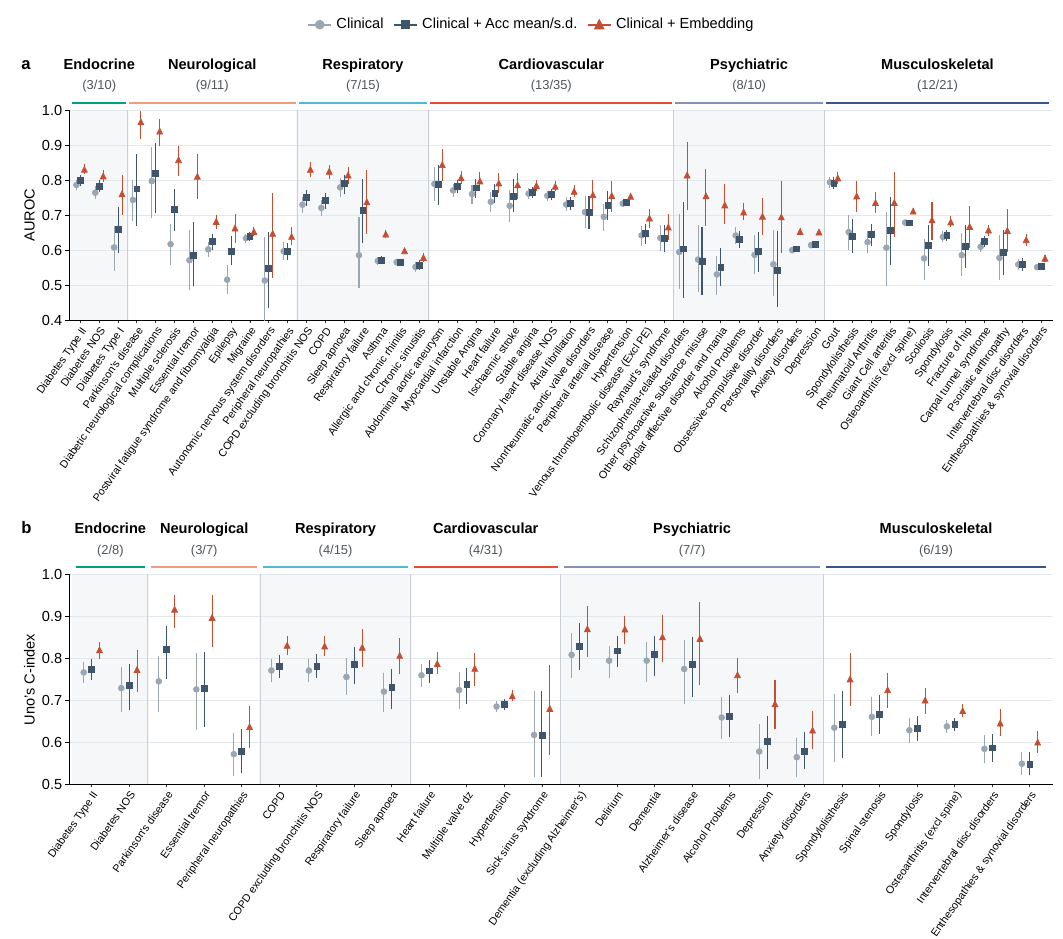}
\caption{\textbf{Performance on prevalent disease classification and incident disease risk prediction.}
\textbf{a}, Prevalent disease classification in UKB, evaluated using AUROC.
\textbf{b}, Six-year incident disease risk prediction in UKB, evaluated using Uno’s C-index.
For both tasks, three models were compared: clinical covariates alone, clinical covariates plus the mean and standard deviation of acceleration (Acc mean/s.d.), and clinical covariates plus fine-tuned Sensori embeddings.
Clinical covariates comprised age, sex, BMI, ethnic background, alcohol consumption and smoking status.
Only conditions for which adding Sensori embeddings significantly improved performance over the clinical covariate model are shown; fractions indicate the number of such conditions among all eligible conditions in each disease category.
Error bars indicate 95\% CIs in the held-out UKB test set.}

\label{fig:disease}
\end{figure}

We further evaluated Sensori alongside models incorporating conventional accelerometry-derived information. 
We first added the mean and standard deviation of acceleration to the clinical covariate model.
For every condition showing a significant improvement relative to the clinical covariate model alone, Sensori retained a higher point estimate (Fig.~\ref{fig:disease}a,b).
Across all eligible conditions, Sensori showed significant improvements in AUROC for 47 of 102 conditions and significant reductions for four in prevalent disease classification.
For incident disease risk prediction, Sensori showed significant improvements in Uno’s C-index for 23 of 87 conditions, with no significant reductions (Supplementary Table~\ref{tab:fig6_paired_bootstrap_summary}).
Among the 81 conditions evaluated in both tasks, Sensori had higher point estimates in both tasks for 19 conditions, in prevalent disease classification only for 25 and in incident disease risk prediction only for three; estimates were not higher for either task in the remaining 34 conditions (Supplementary Fig.~\ref{fig:supp_biomarker_riskfactor_matrix}).

We then compared Sensori with a model combining the clinical covariates with the 18 device-measured behavioural traits shown in Fig.~\ref{fig:2}c.
For prevalent disease classification, Sensori showed significant improvements in AUROC for 47 conditions and significant reductions for two.
For incident disease risk prediction, significant improvements in Uno’s C-index were observed for 22 conditions, with no significant reductions (Supplementary Table~\ref{tab:fig6_paired_bootstrap_summary}).
Condition-specific results for all models are shown in Supplementary Figs.~\ref{fig:fig6a_diagnosis_delta_bar} and~\ref{fig:fig6b_prognosis_delta_bar}.

\section*{Discussion}
To our knowledge, this study is the first to show that a representation learned from 24-hour wrist movement alone can capture diverse dimensions of health and disease and generalise across four population-based cohorts from the United Kingdom, China and the United States. 
Previous studies have linked conventional device-measured behavioural traits, such as time spent at different activity intensities~\cite{stamatakis2022association} and step counts~\cite{master2022association}, to a broad range of chronic diseases and mortality. 
Building on these studies, our work shows that machine-learned movement representations capture information spanning personal characteristics, physical function, prevalent disease and future disease risk. 
Our findings highlight the potential of such representations to provide scalable, objective assessments of behaviour and physical function, reducing reliance on self-report and its associated biases and data-collection burden, and to identify individuals at increased risk of future disease, with implications for earlier intervention and prevention.

Recent systems integrating large language models with wearable and other personal health data, such as Fitbit Coach~\cite{narayanswamy2026towards} and ChatGPT Health~\cite{openai2026health}, reflect growing interest in personalised digital health tools. However, wearable data are typically represented using predefined behavioural summaries that may omit information contained in raw movement signals. Therefore, machine-learned movement representations like Sensori could complement language-model-based health systems by providing more informative representations derived directly from raw wrist movement data. Future work should investigate whether such systems can integrate learned sensor representations as an additional modality, analogous to the integration of visual representations in vision--language models~\cite{zhu2024minigpt}.

For prevalent disease classification, Sensori produced large improvements when added to the clinical baseline, particularly for neurological and psychiatric conditions, including Parkinson’s disease, multiple sclerosis and essential tremor. 
Digital measures such as average acceleration~\cite{redfield2015isosorbide} and, more recently, the 95th percentile of stride velocity~\cite{servais2022stride} have been used in clinical trials to monitor disease progression and treatment response. 
However, developing and validating a sensitive digital biomarker separately for each condition is time-consuming. 
Sensori could provide a general-purpose representation for screening movement signals during endpoint discovery, identifying conditions for which wearable data are likely to provide sensitive measures before disease-specific biomarkers are developed.

Sensori also improved incident disease risk prediction, with the largest gains observed for neurological and psychiatric diseases.
Improvements in other categories were more modest, possibly because behavioural risk factors influence disease through indirect pathways rather than acting as proximal markers of pathology~\cite{diez2014obesity}. 
For disease risk prediction, a recent proteomic study in UKB reported that adding sparse protein signatures to clinical models improved 10-year risk prediction for 67 of 218 eligible diseases, with a median increase in Harrell's C-index of 0.07~\cite{carrasco2024proteomic}. 
In our study, Sensori improved 6-year risk prediction for 26 of 87 conditions, with a median increase in Uno’s C-index of 0.064.
Although these results are not directly comparable, this comparison provides context for the magnitude of Sensori’s gains relative to those achieved with proteomic profiling and highlights wrist movement as a non-invasive, scalable modality for population-level risk stratification.

Several limitations should be considered. 
First, although ELSA and NHANES were designed to provide nationally representative samples, UKB and CKB are susceptible to healthy-volunteer bias. 
Frailer individuals and those with severe disease, multimorbidity or heterogeneous clinical presentations may consequently be underrepresented. 
Evaluation in clinically ascertained populations will be needed to establish generalisability across the full spectrum of health and disease. 
Second, disease analyses were restricted to UKB because comparable cohorts with longitudinal follow-up were unavailable and disease phenotypes are difficult to harmonise across coding systems. 
The recently collected accelerometry data in CKB could provide an important opportunity for external validation once sufficient follow-up has accrued. 
Third, many conditions, particularly neurological diseases, have prodromal periods extending over several years. 
Our analyses may therefore reflect undiagnosed or prodromal disease rather than risk before disease onset~\cite{floud2020body}. 
Landmark analyses progressively excluding cases recorded during the early years of follow-up will be important in future work. 
Finally, Sensori was developed using acceleration signals alone, whereas modern wearable devices can simultaneously capture movement alongside multiple physiological signals. 
Such multimodal wearable measurements were not consistently available at a comparable scale across the cohorts considered here.
Incorporating additional sensor modalities could further enrich the learned representations and capture complementary dimensions of health and disease.

In conclusion, we have demonstrated that 24-hour wrist movement is information-rich and readily obtainable.
Our findings show that learned representations of these signals capture diverse behavioural traits, demographic characteristics, health axes, physical function and disease status. 
When appropriately used and combined with other home-based measurements, wrist movement monitoring could unlock more accessible, precise and preventive care at scale.

\section*{Methods} 

\subsection*{Data sources}
We used raw tri-axial wrist-worn accelerometer data from four population-based cohorts and four human activity recognition benchmark datasets. 
The cohort studies were conducted in the United Kingdom, China and the United States: the UK Biobank (UKB)~\cite{doherty2017large}, the China Kadoorie Biobank (CKB)~\cite{chen2023device}, the English Longitudinal Study of Ageing (ELSA)~\cite{glag197} and the National Health and Nutrition Examination Survey (NHANES)~\cite{cdc_nhanes_pax80g_2026, cdc_nhanes_pax80h_2026}. 
The human activity recognition benchmark datasets were PAMAP2~\cite{reiss2012introducing}, RealWorld~\cite{sztyler2016body}, WISDM~\cite{weiss2019wisdm} and CAPTURE-24~\cite{chan2024capture24}.
All cohort studies followed broadly similar accelerometry protocols. 
Participants wore a research-grade wrist-worn accelerometer continuously for seven consecutive days in UKB, CKB and NHANES and for eight consecutive days in ELSA.

For both UKB and CKB, data were partitioned at the participant level into training (80\%) and test (20\%) sets.
The partitioning strategy differed by cohort.
In UKB, we partitioned geographically to evaluate spatial generalisation.
Participants attending assessment centres south of latitude \(54.50^\circ\) and east of longitude \(-2.98^\circ\) formed the training set, covering most of England.
Participants from the Newcastle and Middlesbrough assessment centres, together with all participants from Scotland and Wales, formed the test set.
CKB was split randomly because assessment-centre information was unavailable.
Data from ELSA and NHANES, together with all human activity recognition benchmark datasets, were reserved for external test sets and did not contribute to pretraining.

The UKB study was approved by the National Health Service (NHS) Research Ethics Service (Ref. 11/NW/0382). 
The CKB movement enhancement was approved by the Chinese Academy of Medical Sciences-Fuwai Hospital/Peking Union Medical College, Ethical Committee (Beijing, China, Ref. 2018-1127) and the Tropical Research Ethics Committee at the University of Oxford (UK, 2019, 18--19).
The ELSA accelerometry enhancement was approved by the South Central Berkshire Research Ethics Committee (Ref. 21/SC/0030). 
The NHANES study was approved by the National Center for Health Statistics Research Ethics Review Board (Protocol \#2011-17).

\subsection*{Accelerometry preprocessing}
Raw tri-axial accelerometer data were processed using actipy~\cite{chan2021actipy}, an open-source Python toolkit developed by the OxWearables group. 
Signals were clipped to \(\pm 3\) g, calibrated to local gravity, low-pass filtered with a cut-off frequency of 5 Hz and downsampled to 10 Hz, balancing predictive performance against computational cost~\cite{yamane2025effects}. 
Days with fewer than 22 hours of wear time, failed calibration, extreme acceleration values or more than ten interruptions were excluded. 
Participant inclusion diagrams for all cohorts are provided in Supplementary Figs.~\ref{fig:supp_ukb_participant_inclusion}--\ref{fig:supp_nhanes_participant_inclusion}.

\subsection*{Sensori network}
Our Sensori architecture was adapted from wav2vec 2.0~\cite{baevski2020wav2vec} for continuous wrist movement signals. 
The model comprised a one-dimensional convolutional feature encoder followed by a transformer backbone, which integrated local movement patterns over time into whole-day behavioural representations.
Each input comprised 24 hours of tri-axial wrist-acceleration signals sampled at 10 Hz.
The convolutional encoder consisted of seven one-dimensional convolutional layers with 512 channels, kernel sizes [10, 3, 3, 3, 2, 2, 2] and strides [5, 2, 2, 2, 2, 2, 2].
Mean pooling reduced the encoded sequence to 288 five-minute embeddings, which were passed to the transformer.

The transformer backbone comprised eight layers with 12 attention heads, a hidden dimension of 768 and learnable relative positional encodings. 
A 768-dimensional day-level embedding was obtained by mean pooling the 288 transformer outputs.
Unless otherwise stated, all results were obtained using the 40-million-parameter model.
We also evaluated smaller (20 million parameters) and larger (80 million parameters) variants and found that the 40-million-parameter model provided the best trade-off between predictive performance and computational efficiency.
Details of the model architecture are specified in Supplementary Table~\ref{tab:supp_model}. 

\subsection*{Pretraining}
We employed two complementary pretraining objectives: masked reconstruction~\cite{devlin2019bert} and participant-level contrastive learning~\cite{oord2018representation}. 
Together, these objectives encouraged the model to capture both within-day temporal structure and stable behavioural characteristics across repeated observations of the same individual.
This combined pretraining strategy was selected following systematic comparison with alternative strategies, including single-task objectives and instance-level contrastive learning (Supplementary Note~\ref{sec:supp_pretraining}).

Masked reconstruction was performed at the five-minute embedding level.
We randomly masked 50\% of the 288 five-minute embeddings after the convolutional encoder and first mean-pooling layer,
before they were passed to the transformer. Mask spans of up to two consecutive embeddings were sampled independently and allowed to overlap.
The model was trained to maximise the similarity between the transformer output and the corresponding target embedding produced by the convolutional encoder at each masked position using the information noise-contrastive estimation (InfoNCE) loss~\cite{oord2018representation}. 
For each masked position, 30 negative embeddings were sampled from the same 24-hour recording.

The contrastive objective was applied to day-level embeddings. 
In each mini-batch, a positive pair was formed for each individual from two randomly selected days; negative pairs were formed from days belonging to different individuals.
Each mini-batch contained 54 individuals. 
Hyperparameters were selected empirically rather than through systematic optimisation owing to computational constraints.

After quality control, the pretraining dataset comprised 502,823 person-days of accelerometer recordings from 91,037 individuals in the UKB and CKB training sets.
The 40-million-parameter Sensori network was pretrained for 300 epochs on three NVIDIA A100 GPUs (80 GB memory each) using the AdamW optimiser~\cite{loshchilov2017decoupled} with a cosine learning rate scheduler and linear warm-up (learning rate, \(4 \times 10^{-5}\)).
Reconstruction loss alone was used for the first 250 epochs, by which point the within-day representations had stabilised.
The contrastive objective was then introduced for the remaining 50 epochs.
This delayed optimisation reduced overfitting arising from the limited number of repeated observations available for each participant.

Three data augmentations were applied during pretraining to increase representation diversity: signal scaling, in which each channel was multiplied by a random scalar sampled from $\mathcal{N}(1,0.1)$; channel swapping, in which the three accelerometer axes were randomly permuted; and random start time, in which each training sample comprised a randomly selected 24-hour window from an individual’s continuous recordings.

\subsection*{Linear probing}
For all analyses other than disease prediction, we evaluated Sensori embeddings using linear probing~\cite{alain2016understanding}: the pretrained model was held fixed, and only a linear model was fitted to its embeddings for each downstream task.
This protocol is widely used in self-supervised representation learning to assess how readily task-relevant information can be recovered from the embeddings without adapting the pretrained model itself.

To ensure comparability, we applied the same linear evaluation framework across all predictor sets, including Sensori embeddings, embeddings from other deep learning models, handcrafted features, and demographic and clinical covariates. 
For binary outcomes, we used class-balanced $L_2$-regularised logistic regression with class weights inversely proportional to class frequency.
For continuous outcomes, we used $L_2$-regularised linear regression (ridge regression).
For incident disease risk prediction, we used $L_2$-regularised Cox proportional hazards models.
Hyperparameters were selected by linear sweep using 20\% of the training set as validation data.

\paragraph*{Minute-level evaluation}
For human activity recognition, Sensori minute-level embeddings were compared with a 22-dimensional handcrafted feature set, MOMENT~\cite{goswami2024moment}, Chronos-2~\cite{ansari2025chronos}, Bio-PM~\cite{tarale2026bio} and Harnet~\cite{yuan2024harnet}.
Linear probes were implemented as class-balanced $L_2$-regularised logistic regression models, with the regularisation hyperparameter $C$ tuned separately for each predictor set.
In each human activity recognition dataset, linear probe performance was estimated by participant-wise five-fold cross-validation, so that all data from a given participant fell within a single fold and no participant contributed to both fitting and evaluation.
Details of data cleaning and implementation are provided in Supplementary Note~\ref{supp_note_har}.

\paragraph*{Day-level evaluation}
For each participant, we $L_2$-normalised and averaged the Sensori 24-hour embeddings across all valid wear days, yielding one participant-level embedding. 
Where predictor values were missing, continuous predictors were imputed using the median and binary predictors using the modal category. 
Continuous predictors were subsequently standardised using the training-partition mean and standard deviation. 
All imputation and standardisation parameters were estimated exclusively from the training partition and applied unchanged to the test data.
Categorical targets were mapped to prespecified binary targets, with the mappings provided in Supplementary Tables~\ref{tab:covariates_ukb}--\ref{tab:covariates_nhanes}. 
Participants missing the relevant target were excluded separately from each analysis, and the resulting sample sizes are reported.

For device-measured behavioural traits, key demographics (age, sex and BMI) and harmonised health measures (smoking status, alcohol drinking status and self-rated health), linear probes were fitted in the UKB training set and applied without refitting to the held-out UKB test set and to CKB, ELSA and NHANES when the corresponding outcome was available.
Unless otherwise stated, 95\% CIs were estimated using 1,000 bootstrap resamples of participants.

For the ten UKB-specific health axes, linear probes were fitted in the UKB training set and evaluated in the held-out UKB test set, with 95\% CIs reported.
ELSA and NHANES physical function items were evaluated separately using five-fold cross-validation, with the mean and standard deviation across the five outer folds reported.

\subsection*{Disease prediction}
Prevalent disease classification and incident disease risk prediction were evaluated in the held-out UKB test set.
Disease outcomes were defined using the pomegranate and CALIBER phenotyping frameworks~\cite{denaxas2019uk,torralbo2025computational}, which integrate primary care records (for a subset of participants), hospital inpatient data and self-reported questionnaire items.
We focused on disease categories clinically relevant to wrist-derived accelerometry: endocrine, neurological, respiratory, musculoskeletal, cardiovascular and psychiatric diseases, covering a total of 123 diseases.

Diagnoses recorded before accelerometer wear were classified as prevalent and those recorded afterwards as incident.
For each incident disease outcome, participants with the corresponding prevalent disease were excluded from the risk set. 
We also excluded events occurring within six months after accelerometer wear to minimise reverse-causation bias. 
Only diseases with at least 60 cases were analysed, yielding 102 diseases for prevalent disease classification and 87 for incident disease risk prediction. 
Detailed disease definitions and case counts for the entire UKB are provided in Supplementary Table~\ref{tab:supp_disease_inclusion}.

For the disease tasks, we applied supervised disease fine-tuning to enrich disease-related signals by aggregating diseases into body organs specified by CALIBER.
Fine-tuning strategy and hyperparameters were selected using participant-disjoint training and validation subsets drawn from the UKB training partition. 
The convolutional feature encoder was frozen, whereas the linear projection, transformer blocks and organ-level readout were updated. 
We then regenerated the day-level embeddings and averaged them across all valid days for each participant to obtain fixed participant-level representations for downstream linear probes (Supplementary Note~\ref{supp:sec_sft}).

Prevalent disease classification was evaluated using logistic regression on the fixed participant-level embeddings. 
Incident disease risk prediction was assessed using Cox proportional hazards models~\cite{cox1972regression} for 6-year incidence after the accelerometer wear between 2013 and 2015, with outcomes censored at death or study end (follow-up censoring dates: March 2023 for England, August 2022 for Scotland, May 2022 for Wales).

We compared four models: a clinical covariate model including age, sex, BMI, ethnic background, smoking status and alcohol consumption; the clinical covariate model with the mean and standard deviation of acceleration; the clinical model with 18 conventional device-measured behavioural traits; and the clinical model with fine-tuned Sensori embeddings. 
Prevalent disease classification was evaluated using AUROC.
Incident disease risk prediction was evaluated using Uno’s C-index~\cite{uno2011c}, which is robust under high levels of censoring. 
The 95\% CIs were estimated by 1,000 bootstrap resamples of participants, and between-model differences were calculated within the same paired resamples. 
Following Carrasco-Zanini et al.~\cite{carrasco2024proteomic}, between-model differences were considered significant when the 95\% CI of the paired differences across bootstrap resamples did not include zero. 
Positive and negative differences meeting this criterion were classified as significant improvements and reductions, respectively.

\section*{Data availability}
The human activity recognition datasets and NHANES data are publicly available from their original sources and the US Centers for Disease Control and Prevention, respectively. ELSA data are available through the UK Data Service. UK Biobank and China Kadoorie Biobank data are available to approved researchers through their respective data-access platforms.

\section*{Code availability}
All code is available at \url{https://github.com/OxWearables/Sensori}.
The pretrained Sensori model weights are available at \url{https://huggingface.co/light156/Sensori}.
A dedicated project website at \url{https://oxwearables.github.io/Sensori/} provides visual explanations and tutorials on using the codebase and pretrained model weights.

\section*{Author contributions}
H.Y. and Y.W. conceived the study, designed the research and drafted the manuscript.
Z.Z., J.C., J.N., Y.S. and G.N. contributed to the machine learning analyses. 
Y.W. and H.Y. performed model pretraining and evaluation.
D.M., K.B., L.B., D.S., C.Y., J.L., M.B., H.L., A.S., D.W.E., L.L., Z.C., N.W., S.D., G.S.C., H.D. and A.D. contributed to clinical interpretation and validation.
A.D. and H.Y. jointly supervised the research. 
All authors reviewed and approved the final manuscript.

\section*{Competing interests}
A.D. is supported by grants from the Wellcome Trust [223100/Z/21/Z], Novo Nordisk, Swiss Re, Health Data Research UK, Google and the British Heart Foundation Centre of Research Excellence [RE/18/3/34214]; has accepted consulting fees from the University of Wisconsin (US National Institutes of Health (NIH) R01 grant)
and Harvard University (NIH R01 grant); received support for presentations or attendance at several conferences; and has received a donation from Swiss Re for accelerometer data collection in the China Kadoorie Biobank.
All other authors declare no competing interests.

\section*{Acknowledgements} 
We thank all participants in UKB, CKB, ELSA and NHANES for their generous contribution to the original data collection efforts.
This work uses data provided by patients and collected by the NHS as part of their care and support.
This research has been conducted using the UK Biobank Resource under Application Number 59070.
We thank the Biomedical Research Computing (BMRC) team at the University of Oxford for assistance with configuring the computing resources used in this study.
For the purpose of open access, the author(s) has applied a Creative Commons Attribution (CC BY) licence to any Author Accepted Manuscript version arising.

A.D.’s research team is supported by a range of grants from the Wellcome Trust [223100/Z/21/Z, 227093/Z/23/Z], GSK, Boehringer Ingelheim, Google, National Institutes of Health’s Oxford Cambridge Scholars Program, EPSRC Centre for Doctoral Training in Health Data Science (EP/S02428X/1), British Heart Foundation Centre of Research Excellence (grant number RE/18/3/34214), Cancer Research UK, and funding administered by the Danish National Research Foundation in support of the Pioneer Centre for SMARTbiomed.
H.Y. and K.B. acknowledge fellowship support from the Nuffield Department of Population Health.
H.Y. and L.B. also acknowledge support from the Wellcome Trust [223100/Z/21/Z].
Y.W. acknowledges funding from the Pioneer Centre for SMARTbiomed.
G.S.C. is a National Institute for Health and Care Research (NIHR) Senior Investigator. The views expressed in this article are those of the author(s) and not necessarily those of the NIHR or the Department of Health and Social Care.
The China Kadoorie Biobank receives funding from the National Natural Science Foundation of China (82192901, 82192904, 82192900, 82388102) and the Noncommunicable Chronic Diseases-National Science and Technology Major Project (2023ZD0510101, 2023ZD0510100).

\bibliographystyle{unsrtnat}
\bibliography{sn-bibliography}

\clearpage

\begin{table}[H]
\centering
\caption{\textbf{Characteristics of the study populations.} Data are shown for eligible participants from four population-based cohorts: UKB, CKB, ELSA and NHANES.}
\label{tab:population_characteristics}
\begin{threeparttable}
\fontsize{8.5pt}{9.4pt}\selectfont
\setlength{\tabcolsep}{3pt}
\renewcommand{\arraystretch}{1.0}
\begin{tabularx}{\textwidth}{@{}>{\raggedright\arraybackslash}p{0.32\textwidth}*{4}{>{\centering\arraybackslash}X}@{}}
\toprule
\textbf{Characteristic} & \makecell{\textbf{UKB} \\ \scriptsize $n = \text{94,089}$} & \makecell{\textbf{CKB} \\ \scriptsize $n = \text{20,626}$} & \makecell{\textbf{ELSA} \\ \scriptsize $n = \text{3,320}$} & \makecell{\textbf{NHANES} \\ \scriptsize $n = \text{4,605}$} \\
\midrule
\addlinespace[2pt]
\multicolumn{5}{@{}l}{\textit{\textbf{Study design}}} \\[1pt]
Device & Axivity AX3 & Axivity AX3 & Axivity AX3 & ActiGraph GT3X+ \\
Wrist placement & Dominant wrist & Dominant wrist & Dominant wrist & Non-dominant wrist \\
Country & UK & China & UK & USA \\
\addlinespace[2pt]
\multicolumn{5}{@{}l}{\textit{\textbf{Demographics}}} \\[1pt]
Age (years), median {[}P25--P75{]} & 63.6 {[}56.4--68.7{]} & 65.5 {[}57.8--71.7{]} & 69.0 {[}61.5--75.9{]} & 59.0 {[}51.0--67.0{]} \\
\quad 33.6--49.9 & 7,675 (8.2\%) & 375 (1.8\%) & 28 (0.8\%) & 963 (20.9\%) \\
\quad 50--59 & 26,757 (28.4\%) & 6,404 (31.1\%) & 656 (19.8\%) & 1,345 (29.2\%) \\
\quad 60--69 & 41,836 (44.5\%) & 7,303 (35.4\%) & 1,104 (33.3\%) & 1,435 (31.2\%) \\
\quad 70.0--98.9 & 17,821 (18.9\%) & 6,526 (31.7\%) & 1,532 (46.1\%) & 862 (18.7\%) \\
Female sex, $n$ (\%) & 52,984 (56.3\%) & 13,446 (65.2\%) & 1,847 (55.6\%) & 2,401 (52.1\%) \\
BMI (kg/m$^{2}$), median {[}P25--P75{]} & 26.0 {[}23.6--29.0{]} & 24.3 {[}22.2--26.7{]} & 27.3 {[}24.4--31.1{]} & 28.6 {[}24.9--33.2{]} \\
\quad Underweight (10.4--18.4) & 488 (0.5\%) & 617 (3.0\%) & 18 (0.7\%) & 56 (1.2\%) \\
\quad Normal weight (18.5--24.9) & 35,407 (38.2\%) & 11,098 (53.9\%) & 788 (30.0\%) & 1,112 (24.2\%) \\
\quad Overweight (25.0--29.9) & 38,625 (41.6\%) & 7,541 (36.6\%) & 1,007 (38.3\%) & 1,529 (33.2\%) \\
\quad Obese (30.0--82.9) & 18,247 (19.7\%) & 1,340 (6.5\%) & 817 (31.1\%) & 1,905 (41.4\%) \\
\addlinespace[2pt]
\multicolumn{5}{@{}l}{\textit{\textbf{Device metrics}}} \\[1pt]
Acceleration (mg), median {[}P25--P75{]} & 27.1 {[}22.5--32.6{]} & 29.6 {[}22.5--37.9{]} & 21.7 {[}17.1--27.3{]} & 22.6 {[}17.4--28.3{]} \\
\quad 2.5--19.9 & 13,427 (14.3\%) & 3,598 (17.4\%) & 1,364 (41.1\%) & 1,716 (37.3\%) \\
\quad 20--30 & 47,208 (50.2\%) & 6,959 (33.7\%) & 1,406 (42.4\%) & 1,965 (42.7\%) \\
\quad 30--40 & 26,136 (27.8\%) & 5,852 (28.4\%) & 450 (13.6\%) & 730 (15.9\%) \\
\quad 40.0--256.2 & 7,317 (7.8\%) & 4,217 (20.4\%) & 99 (3.0\%) & 194 (4.2\%) \\
Steps per day, median {[}P25--P75{]} & \makecell{7,938 \\ {[}6,007--10,126{]}} & \makecell{10,105 \\ {[}7,306--13,165{]}} & \makecell{5,440 \\ {[}3,446--7,648{]}} & \makecell{6,639 \\ {[}4,416--9,236{]}} \\
\quad 0--4,999 & 13,700 (14.6\%) & 2,160 (10.5\%) & 1,471 (44.3\%) & 1,451 (31.5\%) \\
\quad 5,000--7,499 & 27,780 (29.5\%) & 3,307 (16.0\%) & 977 (29.4\%) & 1,248 (27.1\%) \\
\quad 7,500--9,999 & 27,998 (29.8\%) & 4,666 (22.6\%) & 536 (16.1\%) & 1,016 (22.1\%) \\
\quad 10,000--37,288 & 24,611 (26.2\%) & 10,493 (50.9\%) & 336 (10.1\%) & 890 (19.3\%) \\
Sleep duration (h), median {[}P25--P75{]}\textsuperscript{\textdagger} & 6.86 {[}6.24--7.42{]} & 6.03 {[}5.23--6.80{]} & 6.77 {[}6.08--7.41{]} & \textemdash \\
\quad 0.06--5.99 & 17,004 (18.1\%) & 10,036 (48.9\%) & 665 (23.0\%) & \textemdash \\
\quad 6--8 & 69,057 (73.5\%) & 9,450 (46.0\%) & 1,948 (67.4\%) & \textemdash \\
\quad 8.00--17.49 & 7,849 (8.4\%) & 1,045 (5.1\%) & 279 (9.6\%) & \textemdash \\
\addlinespace[2pt]
\bottomrule
\end{tabularx}
\begin{tablenotes}[flushleft]
\fontsize{8.3pt}{9.2pt}\selectfont
\item Values are median [P25--P75] or \textit{n} (\%); P25 and P75 denote the 25th and 75th percentiles, respectively. Percentages use variable-specific non-missing denominators.
\item BMI was taken from the baseline assessment visit for UKB. For CKB, ELSA and NHANES, BMI was taken from the same resurvey as the accelerometer data.
\item[\textdagger] Sleep duration was not available for NHANES; missing values are shown as \textemdash.
\end{tablenotes}
\end{threeparttable}
\end{table}

\clearpage
\section*{Figures}

\clearpage

\clearpage


\section*{Supplementary material}

# Supplementary Materials: Learning Human Health and Diseases from 24-hour Wrist Movement

## Contents



## List of Tables



## List of Figures



# 1 Sensori model development

## 1.1 Model architecture

The Sensori model consists of a convolutional feature encoder followed by a transformer encoder. The convolutional feature encoder transforms a 24-hour sequence of raw movement signals $\mathcal{X}$ into minute-level embeddings, which are aggregated by mean pooling into 288 five-minute-level embeddings $\mathcal{Z} : z_1, \ldots, z_{288}$. These embeddings are fed into the transformer encoder, which produces contextualised representations $\mathcal{H} : h_1, \ldots, h_{288}$ capturing dependencies across the full 24-hour period. A day-level representation $h_{day}$ is obtained by mean pooling the transformer output $\mathcal{H}$. Two configurations of the Sensori model were used: a 20M-parameter model and a 40M-parameter model. Supplementary Table 1 summarises their architecture in full.

Supplementary Table 1: **Sensori model architecture.**

| **Component** | **20M model** | **40M model** |
|---|---|---|
| ***Feature encoder (1-D convolutional stack)*** | | |
| Input channels | 3 | 3 |
| Convolutional layers | 7 | 7 |
| Channels per layer | 256 | 512 |
| Kernel sizes (layers 1-7) | 10, 3, 3, 3, 2, 2, 2 | 10, 3, 3, 3, 2, 2, 2 |
| Strides (layers 1-7) | 5, 2, 2, 2, 2, 2, 2 | 5, 2, 2, 2, 2, 2, 2 |
| Normalisation | GroupNorm (layer 1 only) | GroupNorm (layer 1 only) |
| Parameters | 991,232 | 3,948,544 |
| ***Transformer encoder*** | | |
| Input projection (conv to embedding) | Linear(256, 512) | Linear(512, 768) |
| Embedding dimension | 512 | 768 |
| Encoder layers | 8 | 8 |
| Attention heads | 8 | 12 |
| Head dimension | 64 | 64 |
| Feed-forward dimension | 1024 | 1024 |
| Positional embedding | Absolute (fixed sinusoidal) | Relative (conv, kernel 32, groups 16) |
| Parameters | 16,823,296 | 32,702,752 |
| ***Self-supervised projection heads*** | | |
| Masked / rank projection dim. (z) | 512 | 768 |
| Contrastive embedding dim. | 512 | 512 |
| Parameters (mask token + all heads) | 1,864,754 | 3,976,498 |
| ***Total*** | | |
| Total parameters | 19,811,890 | 41,023,314 |
| Total parameters (millions) | 19.8M | 41.0M |

## 1.2 Pretraining objectives

We explored three pretraining objectives: masked reconstruction, instance contrastive, and day contrastive. Wrist movement is a rich signal with health information encoded at multiple temporal scales; we therefore hypothesised that pretraining with multiple objectives spanning different temporal scales would capture complementary information in wrist movement.

### 1.2.1 Masked reconstruction of five-minute-level embeddings

Similar to masked language modelling [1] and wav2vec2 in audio [2], we masked a random subset of the five-minute-level embeddings and trained the model to reconstruct the masked embeddings from the unmasked ones. We reasoned that reconstructing latent embeddings would induce a better inductive bias than reconstructing the raw waveform directly.

At every masked position $t$, the transformer output $h_t$ was trained to identify the true embedding $z_t$ among $K$ distractor embeddings, sampled uniformly from the other masked positions of the same 24-hour recording. The

loss is defined as:

$$\mathcal{L}_{mask} = -\sum_{t \in \mathcal{M}} \log \frac{\exp(\text{sim}(h_t, z_t)/\tau)}{\sum_{k=1}^{K+1} \exp(\text{sim}(h_t, z_k)/\tau)} \tag{1}$$

where $\mathcal{M}$ is the set of masked positions, the sum in the denominator runs over the true embedding $z_t$ and its $K$ distractors, $\tau$ is a temperature hyperparameter, and $\text{sim}(u, v) = u \cdot v/(\|u\|\|v\|)$ is the cosine similarity.

**Weighted masking** As observed in our prior work [3], when using free-living movement data for pretraining, weighted sampling of windows by activity intensity was essential to reduce ambiguity in the self-supervision signal. Hence, we also tested a variant of masked reconstruction that weighted the sampling of masked positions by the standard deviation of the acceleration signal within each window.

#### 1.2.2 Instance contrastive of five-minute-level embeddings

Instance contrastive was inspired by the observation that, in free-living individuals, movement patterns are confounded by numerous factors such as environment and life events. The goal of instance contrastive learning is therefore to isolate individual-specific invariants that persist across these confounders, by directly contrasting matched behavioural states across days, comparing walking with walking, and sleeping with sleeping, rather than walking with sleeping.

Concretely, we first bin all five-minute embeddings into ten deciles of activity intensity. At each training step, we sample a target decile at random, and every participant in the mini-batch contributes one embedding from that decile on each of two different days, giving a positive pair $(z_p, z'_p)$ for participant $p$ that reflects the same activity-intensity level expressed on two different days. Embeddings from the other participants in the mini-batch serve as negatives. The loss is defined as:

$$\mathcal{L}_{inst} = -\sum_{p=1}^{N} \log \frac{\exp(\text{sim}(z_p, z'_p)/\tau)}{\sum_{j=1}^{N} \exp(\text{sim}(z_p, z'_j)/\tau)} \tag{2}$$

where $N$ is the number of participants in the mini-batch, and $\tau$ and $\text{sim}(\cdot, \cdot)$ are as defined in Eq. (1).

#### 1.2.3 Day contrastive

Unlike the two objectives above, day contrastive aims to capture day-level invariants for each participant. Different days of the same participant are treated as a positive pair, while days from other participants serve as negatives. The loss is defined as:

$$\mathcal{L}_{day} = -\sum_{p=1}^{N} \log \frac{\exp(\text{sim}(h^p_{day}, h^{p\prime}_{day})/\tau)}{\sum_{j=1}^{N} \exp(\text{sim}(h^p_{day}, h^{j\prime}_{day})/\tau)} \tag{3}$$

where $h^p_{day}$ and $h^{p\prime}_{day}$ are the day-level representations of two different days for participant $p$, and $N$, $\tau$, and $\text{sim}(\cdot, \cdot)$ are as defined above.

In early experiments we also tried a more causally motivated contrastive approach that grouped participants with similar demographic characteristics (e.g., age, sex, BMI)—common confounders of movement—within the same contrastive batch, but this did not yield a noticeable improvement in performance.

## 1.3 Pretraining experiments

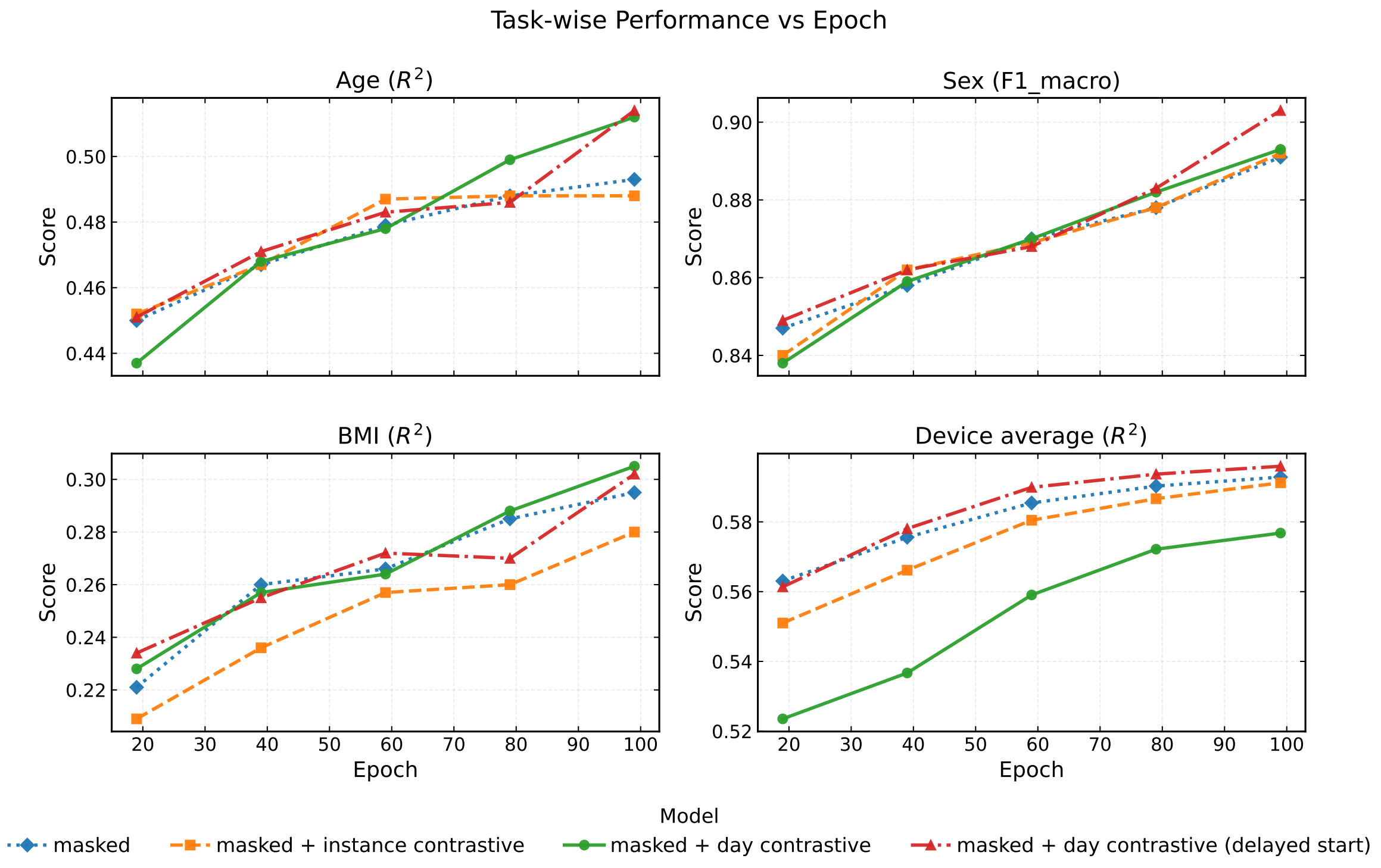


Supplementary Figure 1: **Scaling laws for different pretraining strategies.**

Supplementary Table 2: **Model performance across supervision objectives at epoch 99.** Values are mean ± s.d. across age, sex, BMI, and device outcomes; device average is the mean $R^2$ across device tasks.

| Model | Age ($R^2$) | Sex (F1 macro) | BMI ($R^2$) | Device average ($R^2$) |
|---|---|---|---|---|
| Masked | 0.493 ± 0.011 | 0.891 ± 0.007 | 0.295 ± 0.026 | 0.593 ± 0.016 |
| Masked + instance contrastive | 0.488 ± 0.015 | 0.892 ± 0.006 | 0.280 ± 0.024 | 0.591 ± 0.015 |
| Masked + day contrastive | 0.512 ± 0.021 | 0.893 ± 0.002 | 0.305 ± 0.024 | 0.577 ± 0.015 |
| Masked + day contrastive (delayed start) | 0.514 ± 0.022 | 0.903 ± 0.008 | 0.302 ± 0.023 | 0.596 ± 0.016 |

We define the baseline model for the following ablation as weighted masking off, absolute positional embedding, 20M parameters, and a 5-minute masking temporal horizon with a masking ratio of 0.5. For each ablation, we pretrained on a subset of 10,000 participants in the UK Biobank training set for 100 epochs and performed linear probing on age, sex and BMI, and the core set of algorithmically derived behavioural traits. For multi-objective pretraining, the task weight for masked reconstruction was fixed at 0.9, with the complementary objective weighted at 0.1. Linear probing performance is reported on the same 10,000 participants with five-fold cross-validation. Fig. 1 and Table 2 summarise performance across supervision objectives.

We observed a monotonic increase in performance with training epochs, with the best-performing model, masked + day contrastive (delayed start), achieving the top result on three of four tasks. Although day contrastive could in principle distil more global information into the representations, it was not as effective as expected, possibly because of the limited number of days available for each participant.

Supplementary Table 3: **Ablation study of architecture and masking design choices at epoch 99.** Values are mean ± s.d.; device average is the mean $R^2$ across device tasks.

| Model | Age (R2) | Sex (F1 macro) | BMI (R2) | Device Avg (R2) |
|---|---|---|---|---|
| Masked + day contrastive (delayed start) | 0.514 ± 0.022 | 0.903 ± 0.008 | 0.302 ± 0.023 | 0.596 ± 0.016 |
| + Weighted masking | 0.484 ± 0.023 | 0.895 ± 0.002 | 0.299 ± 0.029 | 0.584 ± 0.013 |
| + Model size 40M parameters | 0.554 ± 0.010 | 0.930 ± 0.008 | 0.375 ± 0.024 | 0.615 ± 0.014 |
| + Relative positional embedding | 0.515 ± 0.015 | 0.912 ± 0.007 | 0.330 ± 0.022 | 0.588 ± 0.016 |
| + Masking temporal horizon: 10 mins | 0.488 ± 0.012 | 0.884 ± 0.005 | 0.284 ± 0.029 | 0.568 ± 0.019 |
| + Masking temporal horizon: 30 mins | 0.465 ± 0.016 | 0.864 ± 0.008 | 0.265 ± 0.024 | 0.562 ± 0.016 |

Finally, Table 3 informed the other key design choices for Sensori, namely the use of the 40M model, relative positional embedding, a masking temporal horizon of 5 minutes, and pretraining without weighted masking for the masked reconstruction task.

# 2 Study populations and benchmark datasets

## 2.1 Prospective cohort studies

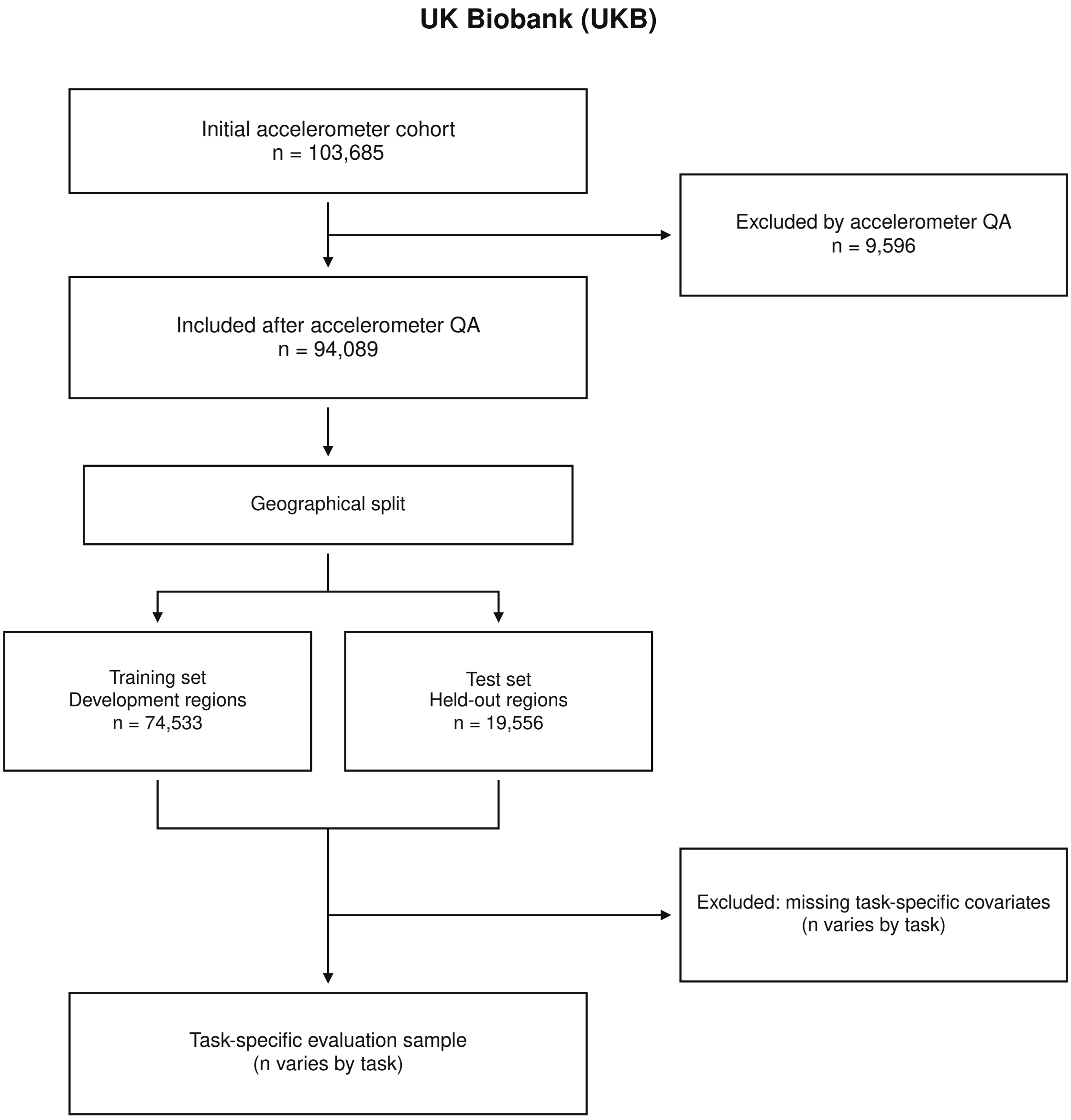


Supplementary Figure 2: **Participant inclusion for UKB.**

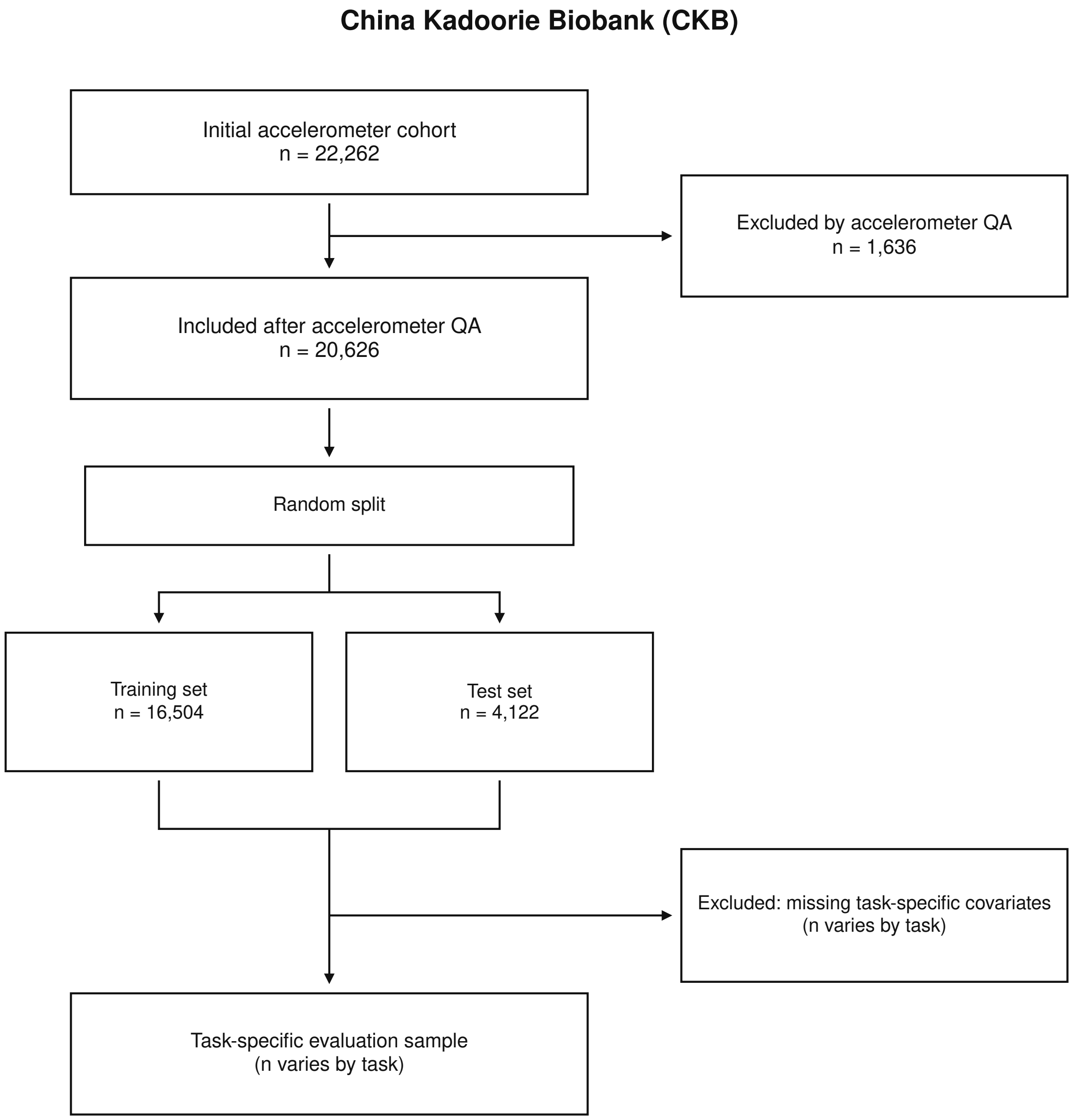


Supplementary Figure 3: **Participant inclusion for CKB.**

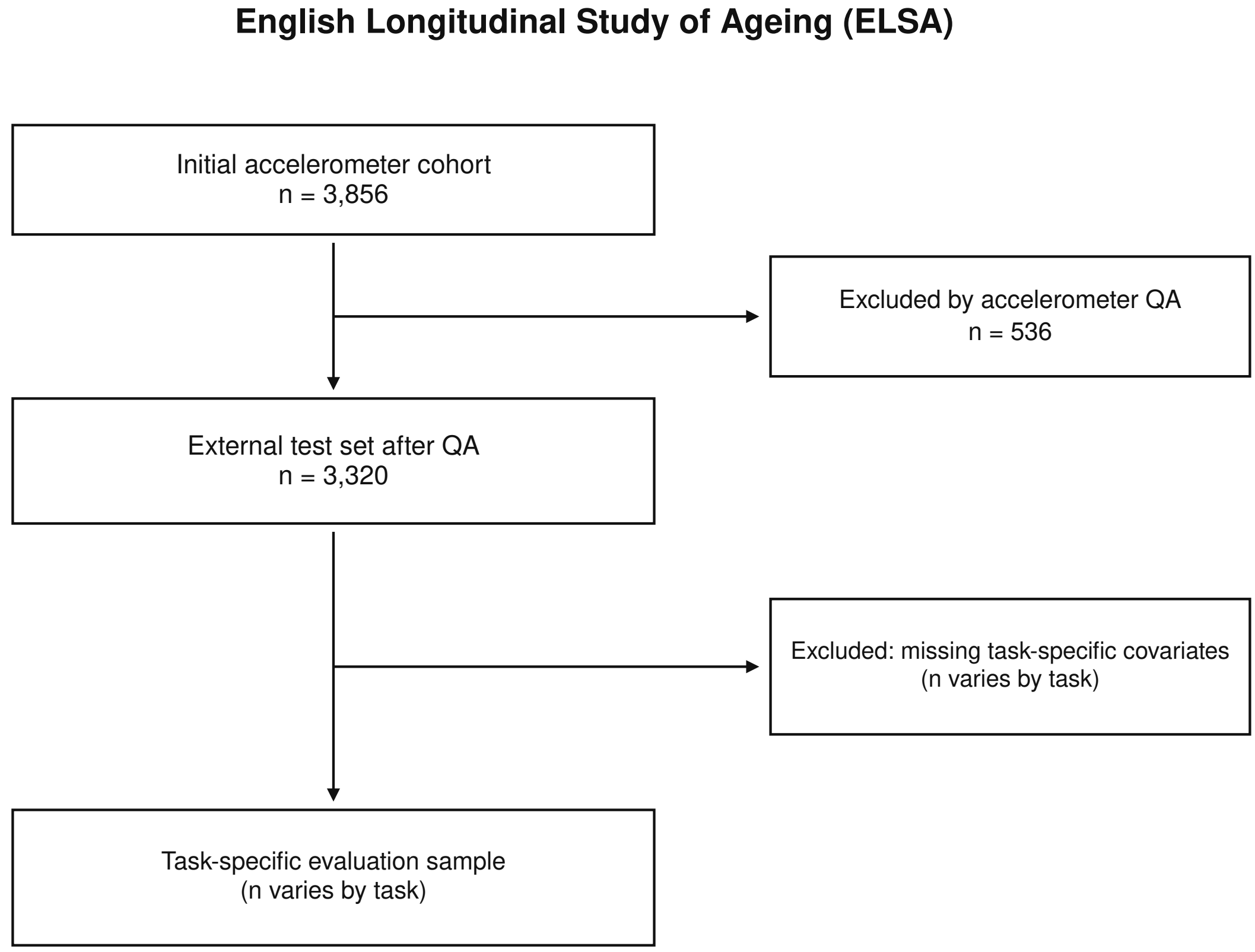


Supplementary Figure 4: **Participant inclusion for ELSA.**

## National Health and Nutrition Examination Survey (NHANES)

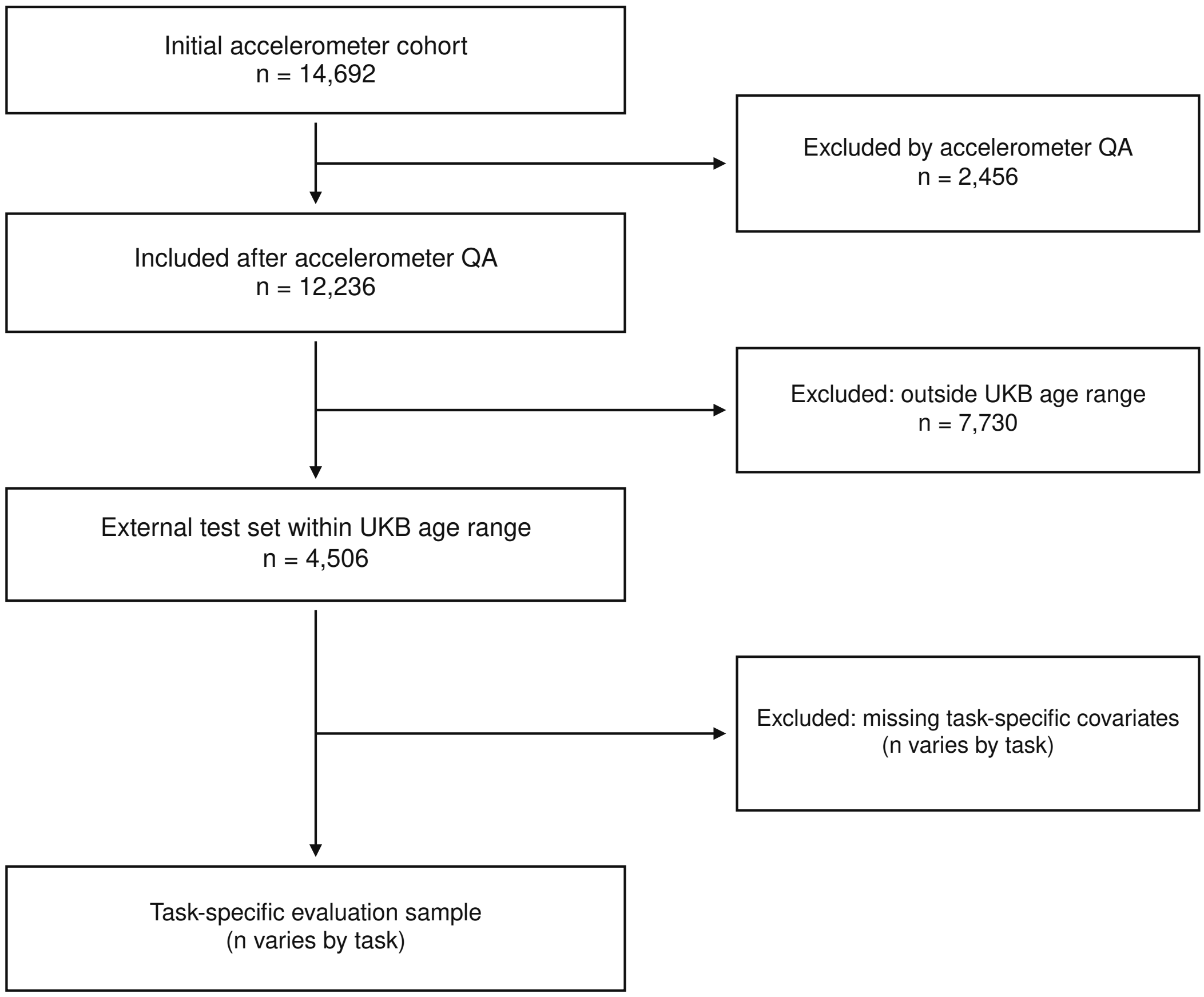


Supplementary Figure 5: **Participant inclusion for NHANES.**

Supplementary Table 4: **UKB variable definitions.** Demographic covariates, harmonised health axes and UKB-specific health axes used in the analyses. Unmapped non-response categories were treated as missing.

| Name | Field ID | Analysis coding |
|---|---|---|
| ***Demographic covariates*** | | |
| Sex | 31 | Female = 0; Male = 1 |
| Age | 34, 52, 90011 | Continuous (years); derived from year and month of birth and accelerometer wear end date |
| BMI | 23104 | Continuous (kg $m^{-2}$) |
| Ethnic background | 21000 | White = 1; all other ethnic backgrounds = 0 |
| ***Harmonised health axes*** | | |
| Smoking status | 1239 | No = 0; Yes, on most or all days, or only occasionally = 1 |
| Regular alcohol drinking | 1558 | Never or only on special occasions = 0; One to three times a month or more often = 1 |
| Self-rated health | 2178 | Fair or poor = 0; Good or excellent = 1 |
| ***UKB-specific health axes*** | | |
| Long-standing illness | 2188 | No or do not know = 0; Yes = 1 |
| Getting up easily | 1170 | Not very easy or not at all easy = 0; Fairly easy or very easy = 1 |
| Nap during day | 1190 | Never/rarely = 0; Sometimes or usually = 1 |
| Sleeplessness | 1200 | Never/rarely = 0; Sometimes or usually = 1 |
| Daytime dozing | 1220 | Never/rarely = 0; Sometimes, often or all of the time = 1 |
| Mood swings | 1920 | No = 0; Yes = 1 |
| Tiredness/lethargy | 2080 | Not at all = 0; Several days, more than half the days, or nearly every day = 1 |
| Brisk walking pace | 924 | Slow pace or steady average pace = 0; Brisk pace = 1 |
| Shortness of breath | 4717 | No = 0; Yes = 1 |
| Happiness | 4526 | Extremely/very/moderately unhappy = 0; Extremely/very/moderately happy = 1 |

Supplementary Table 5: **CKB variable definitions.** Demographic covariates were harmonised to the labels and coding used across cohorts.

| Name | CKB variable | Analysis coding |
|---|---|---|
| Sex | `is_female` | Female = 0; Male = 1 |
| Age | `age_at_study_date_x100` | Continuous (years); divided by 100 |
| BMI | `bmi_calc` | Continuous (kg $m^{-2}$) |

Supplementary Table 6: **ELSA variable definitions.** Unless otherwise stated, variables were obtained from ELSA wave 10. Unmapped non-response categories were treated as missing.

| Name | ELSA variable | Analysis coding |
| --- | --- | --- |
| ***Demographic covariates*** | | |
| Sex | `indsex` | 1 (Male) = 1; 2 (Female) = 0 |
| Age | `indobyr` | Continuous (years); calculated from year of birth and accelerometer start date, with waves 1 and 3 used when wave-10 birth-year values were missing |
| BMI | `height`, `weight`, `estwt`; `bmival`, `r8mbmi` | Continuous (kg $m^{-2}$); calculated from measured height and weight, with estimated weight (`estwt`) used when measured weight was unavailable; `bmival` from wave 11 and `r8mbmi` from wave 8 were used when wave-10 BMI values were missing |
| ***Harmonised health axes*** | | |
| Smoking status | `hesmk`, `heska` | `hesmk` (ever smoked) = 2 (No) or `heska` (currently smokes) = 2 (No) mapped to 0; `heska` = 1 (Yes) mapped to 1 |
| Regular alcohol drinking | `scako` | 6–8 (Once every couple of months or less often) mapped to 0; 1–5 (Once or twice a month or more often) mapped to 1 |
| Self-rated health | `hehelf` | 1 (Excellent), 2 (Very good) or 3 (Good) mapped to 1; 4 (Fair) or 5 (Poor) mapped to 0 |
| ***ELSA-specific physical function items*** | | |
| Participants were asked whether, because of a health problem, they had difficulty doing any of the listed activities. | | |
| Walk 100 yards | `hemobwa` | 0 (Not mentioned) = 0; 1 (Mentioned) = 1 |
| Sit 2 hours | `hemobsi` | |
| Stand from chair | `hemobch` | |
| Climb several flights | `hemobcs` | |
| Climb one flight | `hemobcl` | |
| Stoop/crouch/kneel | `hemobst` | |
| Reach above shoulder | `hemobre` | |
| Push/pull objects | `hemobpu` | |
| Lift/carry > 10 lb | `hemobli` | |
| Pick up 5p coin | `hemobpi` | |

Supplementary Table 7: **NHANES variable definitions.** Variables were obtained from the 2011–2012 and 2013–2014 survey cycles and had identical names and coding in both cycles. Unmapped non-response categories were treated as missing.

| Name | NHANES variable | Analysis coding |
|---|---|---|
| ***Demographic covariates*** | | |
| Sex | RIAGENDR | 1 (Male) = 1; 2 (Female) = 0 |
| Age | RIDAGEYR | Continuous (years) |
| BMI | BMXBMI; WHD010, WHD020 | Continuous (kg m$^{-2}$); measured BMI (BMXBMI) was used where available, with BMI calculated from self-reported height (WHD010) and weight (WHD020) otherwise |
| ***Harmonised health axes*** | | |
| Smoking status | SMQ020, SMQ040 | SMQ020 (smoked at least 100 cigarettes) = 2 (No) or SMQ040 (currently smokes) = 3 (Not at all) mapped to 0; SMQ040 = 1 (Every day) or 2 (Some days) mapped to 1 |
| Regular alcohol drinking | ALQ101, ALQ120Q, ALQ120U | Drinking frequency was converted to days per year using ALQ120Q and ALQ120U; fewer than 12 days per year, or ALQ101 = 2 (No) when frequency was unavailable, mapped to 0; at least 12 days per year mapped to 1 |
| Self-rated health | HSD010 | 1 (Excellent), 2 (Very good) or 3 (Good) mapped to 1; 4 (Fair) or 5 (Poor) mapped to 0 |
| ***NHANES-specific physical function items*** | | |
| Participants were asked how much difficulty they had doing each activity by themselves and without using special equipment. | | |
| Walk 1/4 mile | PFQ061B | 1 (No difficulty), or responses skipped at PFQ059A after negative screening (ages 20–59), mapped to 0; 2 (Some difficulty), 3 (Much difficulty) or 4 (Unable to do) mapped to 1; skipped PFQ061B/C responses when PFQ054 = 1 also mapped to 1 |
| Walk 10 steps | PFQ061C | |
| Stoop/crouch/kneel | PFQ061D | |
| Lift/carry | PFQ061E | |
| House chores | PFQ061F | |
| Prepare meals | PFQ061G | |
| Walk between rooms | PFQ061H | |
| Stand from chair | PFQ061I | |
| Get in/out of bed | PFQ061J | |
| Eat with utensils | PFQ061K | |
| Dress self | PFQ061L | |
| Stand long periods | PFQ061M | |
| Sit long periods | PFQ061N | |
| Reach overhead | PFQ061O | |
| Grasp small objects | PFQ061P | |
| Go out to events | PFQ061Q | |
| Attend social events | PFQ061R | |
| Leisure at home | PFQ061S | |
| Push/pull objects | PFQ061T | |

## 2.2 Datasets for Human Activity Recognition

Supplementary Table 8: **Overview of the human activity recognition datasets used in this study.** Subject and class counts and analysed durations refer to the curated analysis subsets rather than the complete source datasets.

| Dataset | # Subjects | # Classes | Total wear time (h) | Raw Hz | Device | Reference |
|---|---|---|---|---|---|---|
| PAMAP2 | 8 | 8 | 4.5 | 100 | Colibri IMU | [4] |
| RealWorld | 15 | 8 | 17.9 | 50 | LG G Watch R | [5] |
| WISDM | 51 | 18 | 38.3 | 20 | LG G Watch | [6] |
| CAPTURE-24 | 151 | 4 | 2444.1 | 100 | Axivity AX3 | [7] |

Supplementary Table 9: **Class composition of the human activity recognition datasets used in this study.** Counts indicate the number of 1-minute windows included in the analysis for each class; percentages are calculated within each dataset.

| Dataset | Activity class | Windows, $n$ | Proportion (%) |
|---|---|---|---|
| PAMAP2 | lying down | 64 | 13.2 |
| | sitting | 63 | 13.0 |
| | standing | 64 | 13.2 |
| | ironing | 79 | 16.3 |
| | vacuum cleaning | 61 | 12.6 |
| | walking | 78 | 16.0 |
| | ascending stairs | 41 | 8.4 |
| | descending stairs | 36 | 7.4 |
| RealWorld | lying down | 295 | 14.6 |
| | sitting | 295 | 14.6 |
| | standing | 292 | 14.5 |
| | walking | 301 | 14.9 |
| | climbing up | 262 | 13.0 |
| | climbing down | 219 | 10.8 |
| | running | 327 | 16.2 |
| | jumping | 28 | 1.4 |
| WISDM | sitting | 206 | 5.6 |
| | typing | 203 | 5.5 |
| | writing | 204 | 5.5 |
| | eating pasta | 200 | 5.4 |
| | eating sandwich | 197 | 5.4 |
| | eating chips | 208 | 5.6 |
| | eating soup | 203 | 5.5 |
| | drinking from cup | 201 | 5.5 |
| | standing | 204 | 5.5 |
| | brushing teeth | 210 | 5.7 |
| | folding clothes | 205 | 5.6 |
| | clapping | 208 | 5.6 |
| | walking | 209 | 5.7 |
| | stairs | 201 | 5.5 |
| | jogging | 206 | 5.6 |
| | dribbling basketball | 206 | 5.6 |
| | kicking soccer ball | 209 | 5.7 |
| | catch tennis ball | 202 | 5.5 |
| CAPTURE-24 | sleep | 113,067 | 40.1 |
| | sedentary | 111,248 | 39.4 |
| | light | 44,670 | 15.8 |
| | moderate-vigorous | 13,062 | 4.6 |

# 3 Evaluations

## 3.1 Human activity recognition benchmarks

We processed each human activity recognition (HAR) dataset using the dataset-specific parsing and quality-control procedures described by Yuan et al. [8]. The only change was that we segmented the recordings into 60-s windows with a 30-s stride. After data-quality filtering, the dataset characteristics and class distributions used in this study are reported in Tables 8 and 9, respectively.

We compared Sensori with five methods: handcrafted features, MOMENT [9], Chronos-2 [10], Bio-PM [11], and Harnet [8]. For handcrafted features, MOMENT, Chronos-2, and Sensori, we applied a shared preprocessing pipeline. At each dataset's native sampling frequency, the tri-axial accelerometer signal was low-pass filtered using an eighth-order Butterworth filter with a cutoff frequency of 5 Hz and downsampled to 10 Hz, leading to 600 time points per input window. For Bio-PM and Harnet, we used their publicly released preprocessing pipelines and pretrained checkpoints. All deep learning models were kept frozen during embedding extraction.

**Handcrafted features.** Following Yuan et al. [8], we calculated 22 features from each 60-s window. These comprised the mean, standard deviation, and range of each accelerometer axis (nine features); the pairwise Pearson correlations between the $x$-, $y$-, and $z$-axes (three features); and statistics of the acceleration magnitude, $m_t = \sqrt{x_t^2 + y_t^2 + z_t^2}$. The magnitude-based statistics included its mean, standard deviation, range, median absolute deviation, mean absolute deviation from 1 g, skewness, and kurtosis (seven features). We additionally calculated the spectral entropy of the magnitude signal and the frequencies of the two highest local peaks in its Welch power spectral density estimate (three features), yielding 22 features in total.

**MOMENT.** MOMENT [9] is a general-purpose pretrained foundation model for time-series analysis. We used the `MOMENT-1-large` model in its embedding mode, which accepts sequences of 512 time points and uses non-overlapping patches of eight samples. Because directly resampling each input window from 600 to 512 time points resulted in inferior performance in our analysis, we instead divided each window into two contiguous, patch-aligned segments of 304 and 296 time points. Each segment was left-padded to a length of 512, with a binary mask indicating the valid positions. The final embeddings were combined using a length-weighted mean, with weights of 304/600 and 296/600.

**Chronos-2.** Chronos-2 [10] is a pretrained foundation model originally designed for time-series forecasting. We passed each tri-axial input window to the encoder as a trivariate series and averaged the resulting patch representations over the observed-signal patches and the three accelerometer axes, producing a 768-dimensional embedding for each window.

**Bio-PM.** For Bio-PM [11], we used the authors' publicly released implementation and model-specific preprocessing pipeline. We used the recommended settings: the checkpoint pretrained with a 50% masking rate and per-axis pooling. This produced a 1,023-dimensional embedding for each input window.

**Harnet.** Harnet [8] is our previously developed human activity recognition model pretrained on UK Biobank. Although Harnet was originally designed for 30-s input windows of 30-Hz tri-axial accelerometer data, its convolutional structure makes it straightforward to adapt to longer inputs. We therefore resampled each window to 30 Hz, fed the full 60-s input into the model in a single pass, and averaged the output over the temporal dimension to obtain a 1,024-dimensional embedding for each input window.

**Sensori.** For Sensori, we extracted the output of the convolutional feature encoder before the projection heads and the transformer, producing a 512-dimensional embedding for each window.

All features and embeddings were evaluated under the same participant-wise five-fold cross-validation protocol, in which all windows from a participant remained within the same fold. Within each outer training fold, we fitted a class-balanced $L_2$-penalised logistic-regression linear probe, selecting the regularisation parameter $C \in \{10^{-4}, 10^{-3}, \ldots, 10^{4}\}$ by participant-wise inner cross-validation. Handcrafted features and Bio-PM embeddings were standardised using statistics calculated exclusively from the corresponding outer training fold. This accounts for the heterogeneous numerical scales of the handcrafted features and follows the released Bio-PM evaluation protocol.

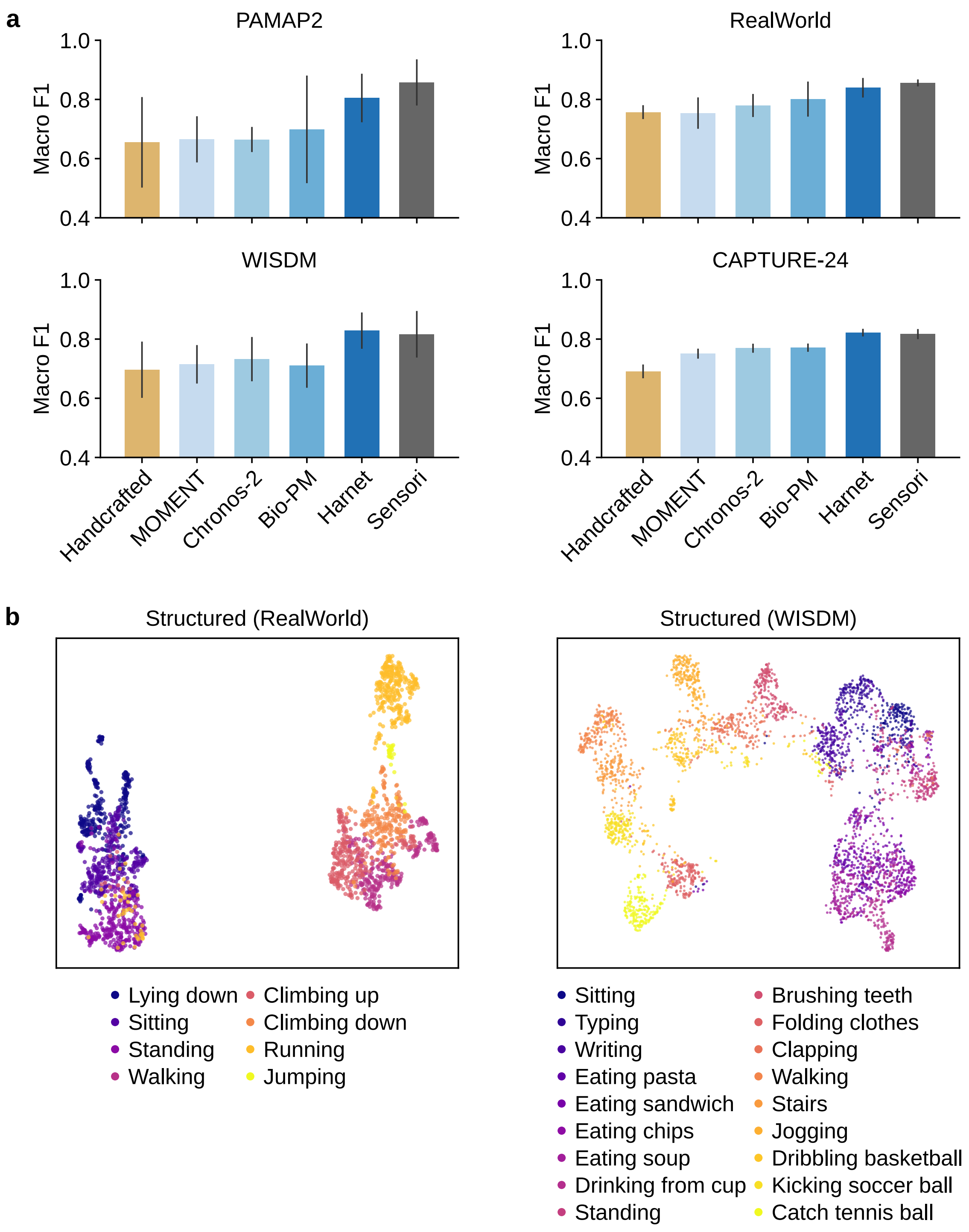


Supplementary Figure 6: **Additional results of human activity recognition benchmarks.** **a**, Macro-averaged F1 scores evaluated on PAMAP2, RealWorld, WISDM and CAPTURE-24. Bars and error bars show the mean and standard deviation across the five participant-grouped outer folds, respectively. **b**, UMAP visualisations of the Sensori embeddings for RealWorld and WISDM, coloured by activity class.

## 3.2 Evaluating day-level behavioural traits

Supplementary Table 10: **Definitions of device-measured behavioural traits.** Large-scale cohort accelerometer data were collected under free-living conditions. To generate the labels for day-level behaviours, we used our previously developed packages [3, 12, 13] to acquire algorithm-derived behavioural traits at the day level. These traits comprised measures of physical activity, steps and sleep.

| Variable | Definition | Units |
|---|---|---|
| ***Physical activity traits*** (biobankAccelerometerAnalysis [12]) | | |
| Acceleration mean | Mean wrist acceleration across all valid wear days. | mg |
| Acceleration s.d. | Standard deviation of wrist acceleration across all valid wear days. | mg |
| Light duration | Mean daily proportion of time in light-intensity activity (activity classifier). | proportion |
| Moderate-vigorous duration | Mean daily proportion of time in moderate-to-vigorous activity. | proportion |
| Sedentary duration | Mean daily proportion of time in sedentary behaviour. | proportion |
| Metabolic equivalent | Mean metabolic equivalent of task (MET), from a per-class lookup applied to the activity classifier output. | METs |
| ***Step traits*** (stepcount [13]) | | |
| Daily step count | Mean daily step count, adjusted for missing wear time. | steps/day |
| Daily walk duration | Mean daily time spent walking, adjusted for missing wear time. | minutes/day |
| Maximum daily cadence | Mean of each day's single highest 1-minute walking cadence, adjusted for missing wear time. | steps/min |
| Mean top-30 cadence | Mean of each day's cadence averaged over its 30 most active walking minutes, adjusted for missing wear time. | steps/min |
| 95th-percentile cadence | Mean of each day's 95th-percentile walking cadence, adjusted for missing wear time. | steps/min |
| ***Sleep traits*** (asleep [3]) | | |
| Sleep onset latency | Time from bedtime to sleep onset. | minutes |
| Total sleep time | All sleep stages combined. | minutes |
| Wake after sleep onset | Wake time occurring after sleep onset. | minutes |
| Sleep efficiency | Total sleep time divided by time in bed. | proportion |
| Total Wake time during sleep | Total wake time from bedtime to rise time. | minutes |
| Total NREM sleep duration | Non-rapid eye movement (NREM) sleep duration (stages N1+N2+N3). | minutes |
| Total REM sleep duration | Rapid eye movement (REM) sleep duration. | minutes |

Supplementary Table 11: **Prediction of device-measured behavioural traits.** Values are Pearson's $r$ with 95% CIs between the observed and predicted trait values. Within each cell, the first line presents the baseline model and the second line, shown in dark blue, presents Sensori. The baseline model included age, sex and BMI. CIs were estimated using 1,000 bootstrap resamples of participants. A dash indicates that the corresponding trait was unavailable.

| Trait | Pearson's $r$ (95% CI) | | | |
|---|---|---|---|---|
| | UKB | CKB | ELSA | NHANES |
| ***Physical activity traits*** | | | | |
| Acceleration mean | 0.370 (0.356–0.384)<br>0.901 (0.878–0.920) | 0.407 (0.383–0.432)<br>0.942 (0.930–0.951) | 0.477 (0.453–0.501)<br>0.939 (0.931–0.946) | 0.439 (0.418–0.458)<br>0.819 (0.807–0.830) |
| Acceleration s.d. | 0.381 (0.370–0.392)<br>0.741 (0.731–0.752) | 0.316 (0.288–0.346)<br>0.814 (0.800–0.827) | 0.469 (0.444–0.494)<br>0.818 (0.794–0.841) | 0.376 (0.327–0.433)<br>0.439 (0.394–0.493) |
| Light duration | 0.292 (0.279–0.305)<br>0.951 (0.949–0.953) | 0.341 (0.315–0.368)<br>0.926 (0.919–0.934) | 0.282 (0.252–0.316)<br>0.945 (0.936–0.952) | 0.223 (0.196–0.252)<br>0.900 (0.895–0.905) |
| Moderate-vigorous duration | 0.325 (0.314–0.338)<br>0.905 (0.899–0.909) | 0.201 (0.170–0.235)<br>0.840 (0.825–0.855) | 0.342 (0.320–0.367)<br>0.900 (0.890–0.908) | 0.323 (0.299–0.347)<br>0.594 (0.568–0.619) |
| Sedentary duration | 0.329 (0.317–0.341)<br>0.939 (0.937–0.941) | 0.292 (0.262–0.322)<br>0.943 (0.938–0.947) | 0.340 (0.308–0.373)<br>0.928 (0.920–0.934) | 0.263 (0.234–0.291)<br>0.836 (0.828–0.845) |
| Metabolic equivalent | 0.251 (0.239–0.263)<br>0.922 (0.919–0.926) | 0.170 (0.136–0.205)<br>0.905 (0.896–0.914) | 0.325 (0.298–0.353)<br>0.923 (0.917–0.929) | 0.317 (0.293–0.342)<br>0.898 (0.891–0.905) |
| ***Step traits*** | | | | |
| Daily step count | 0.281 (0.268–0.292)<br>0.936 (0.934–0.939) | 0.194 (0.158–0.231)<br>0.938 (0.933–0.942) | 0.391 (0.366–0.418)<br>0.952 (0.947–0.956) | 0.399 (0.378–0.421)<br>0.907 (0.901–0.912) |
| Daily walk duration | 0.260 (0.247–0.274)<br>0.940 (0.937–0.942) | 0.108 (0.075–0.145)<br>0.939 (0.935–0.943) | 0.321 (0.292–0.349)<br>0.952 (0.948–0.956) | 0.356 (0.333–0.380)<br>0.939 (0.935–0.942) |
| Maximum daily cadence | 0.351 (0.338–0.363)<br>0.754 (0.746–0.761) | 0.312 (0.282–0.343)<br>0.751 (0.735–0.767) | 0.402 (0.369–0.431)<br>0.797 (0.781–0.813) | 0.385 (0.359–0.411)<br>0.701 (0.686–0.716) |
| Mean top-30 cadence | 0.300 (0.287–0.313)<br>0.815 (0.809–0.820) | 0.247 (0.213–0.278)<br>0.834 (0.823–0.843) | 0.413 (0.380–0.440)<br>0.855 (0.842–0.867) | 0.420 (0.395–0.447)<br>0.789 (0.777–0.800) |
| 95th-percentile cadence | 0.293 (0.280–0.305)<br>0.774 (0.768–0.781) | 0.214 (0.181–0.245)<br>0.779 (0.766–0.791) | 0.384 (0.353–0.413)<br>0.800 (0.786–0.815) | 0.361 (0.334–0.388)<br>0.677 (0.660–0.694) |
| ***Sleep traits*** | | | | |
| Sleep onset latency | 0.081 (0.068–0.095)<br>0.381 (0.365–0.397) | 0.076 (0.044–0.104)<br>0.432 (0.402–0.465) | 0.089 (0.053–0.133)<br>0.397 (0.351–0.442) | – |
| Total sleep time | 0.162 (0.148–0.176)<br>0.866 (0.861–0.871) | 0.107 (0.075–0.136)<br>0.857 (0.848–0.866) | 0.074 (0.036–0.112)<br>0.856 (0.843–0.869) | – |
| Wake after sleep onset | 0.086 (0.072–0.100)<br>0.770 (0.763–0.777) | 0.034 (0.001–0.067)<br>0.747 (0.734–0.761) | 0.031 (-0.005–0.066)<br>0.736 (0.715–0.756) | – |
| Sleep efficiency | 0.120 (0.105–0.134)<br>0.788 (0.781–0.794) | 0.047 (0.015–0.079)<br>0.770 (0.756–0.783) | 0.036 (-0.002–0.072)<br>0.772 (0.754–0.790) | – |
| Total wake time during sleep | 0.089 (0.075–0.104)<br>0.777 (0.770–0.783) | 0.017 (-0.016–0.051)<br>0.765 (0.752–0.779) | 0.024 (-0.011–0.058)<br>0.750 (0.731–0.769) | – |
| Total NREM sleep duration | 0.200 (0.186–0.213)<br>0.792 (0.786–0.798) | 0.150 (0.120–0.181)<br>0.788 (0.776–0.800) | 0.123 (0.086–0.162)<br>0.788 (0.772–0.803) | – |
| Total REM sleep duration | 0.271 (0.258–0.285)<br>0.759 (0.752–0.765) | 0.272 (0.243–0.298)<br>0.647 (0.628–0.666) | 0.302 (0.269–0.334)<br>0.770 (0.753–0.785) | – |

## 3.3 Evaluating key demographic characteristics and multidimensional health axes

Supplementary Table 12: **Prediction of age, sex and BMI across cohorts.** Within each cell, the first line presents the prediction results using the mean and standard deviation of acceleration, and the second line, shown in dark blue, presents Sensori. Values in parentheses are 95% bootstrap CIs.

| **Cohort** | **Sex (AUROC)** | **Age (Pearson's *r*)** | **BMI (Pearson's *r*)** |
|---|---|---|---|
| UKB | 0.630 (0.622–0.638)<br>0.996 (0.995–0.997) | 0.292 (0.279–0.305)<br>0.837 (0.833–0.841) | 0.273 (0.260–0.285)<br>0.733 (0.725–0.740) |
| CKB | 0.636 (0.616–0.653)<br>0.969 (0.964–0.974) | 0.393 (0.366–0.420)<br>0.765 (0.754–0.776) | 0.092 (0.062–0.122)<br>0.511 (0.481–0.541) |
| ELSA | 0.600 (0.581–0.620)<br>0.988 (0.984–0.991) | 0.427 (0.400–0.453)<br>0.846 (0.836–0.855) | 0.283 (0.249–0.319)<br>0.710 (0.689–0.729) |
| NHANES | 0.535 (0.520–0.551)<br>0.976 (0.972–0.979) | 0.386 (0.351–0.417)<br>0.677 (0.662–0.691) | 0.221 (0.193–0.245)<br>0.596 (0.578–0.614) |

## 3.4 Prediction of multidimensional health axes

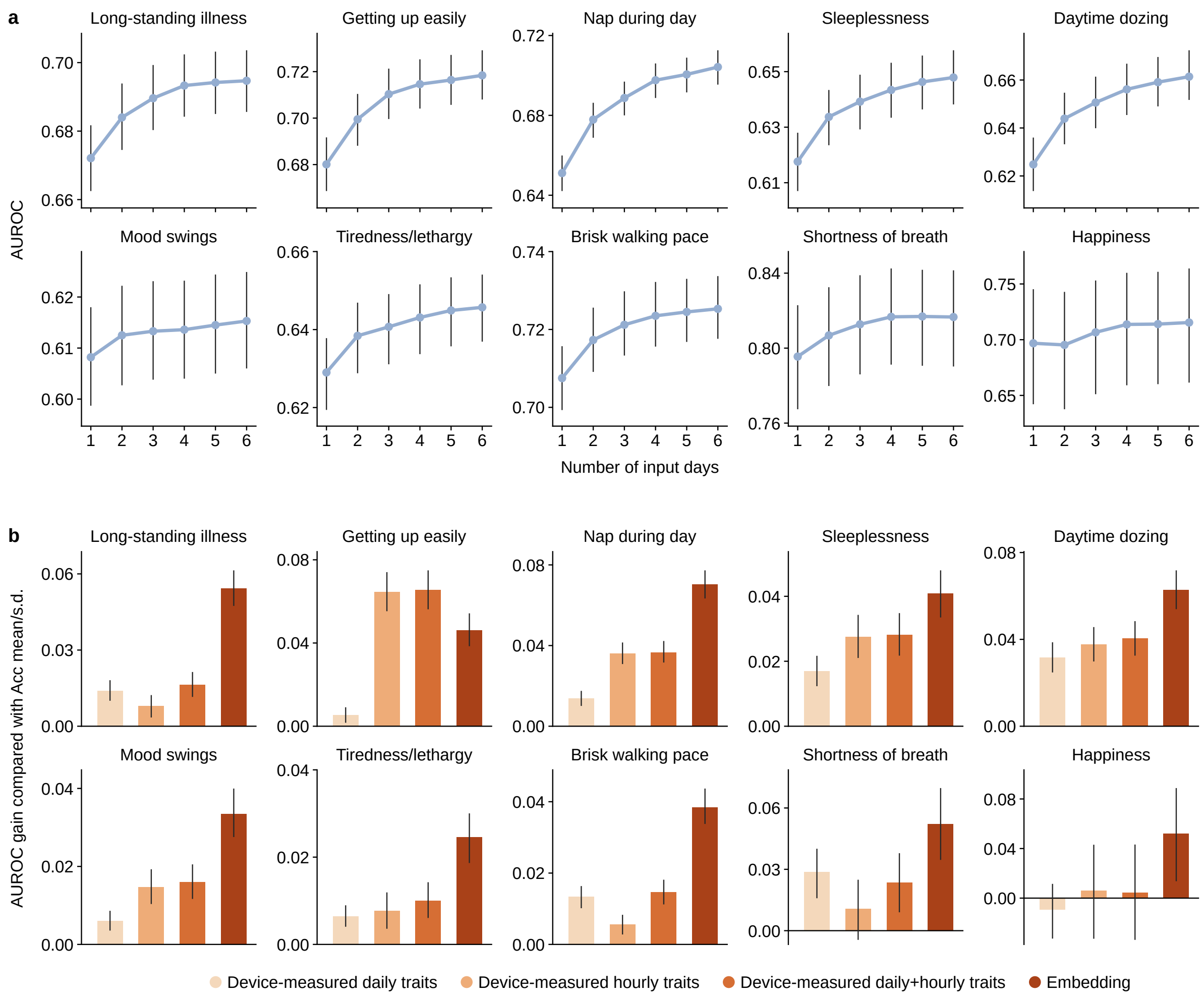


Supplementary Figure 7: **Individual results for the ten UKB health axes.** **a**, Multi-day analysis. **b**, Comparison with conventional device-measured traits. The analyses are identical to those in Fig. 4b and 4c in the main text, with the ten health axes shown separately.

## 3.5 Disease prediction

For disease prediction, we applied supervised fine-tuning to enrich disease-related signals. To select the hyper-parameters and fine-tuning strategy, we used 40% of the UKB training set as a derivation set, split equally into participant-disjoint training and validation sets. To reduce computational cost, we froze the convolutional feature extractor and fine-tuned the linear projection, transformer blocks, and final readout layers.

We evaluated three variants: disease-specific fine-tuning with pretrained weights, disease-specific fine-tuning with randomly initialised weights downstream of the feature encoder, and pretrained shared-organ fine-tuning. The multi-label shared-organ approach was designed to mitigate the extreme class imbalance associated with rare diseases by sharing supervision across diseases with related aetiologies (Table 13). Disease groupings followed CALIBER chapters. This approach also substantially reduced computation: rather than fitting 102 disease-specific models, only one model with six organ-level outputs was required, reducing fine-tuning compute by approximately 99%.

Each day was treated as one observation during training. To prevent participants with more available days from dominating training, each day was assigned weight $w_{ij} = 1/n_i$, where $n_i$ is the number of days available for participant i. After fine-tuning, we regenerated day-level embeddings and averaged them across all available days to obtain participant-level representations for downstream linear probes. We report validation performance for six clinically relevant candidate diseases in Table 14. Shared-organ fine-tuning achieved the best overall performance and was therefore selected for the main analysis. Across all six candidate diseases, disease-specific fine-tuning with pretrained initialisation outperformed the corresponding random-initialisation control; the convolutional feature extractor was frozen in both variants.

Supplementary Table 13: **Disease-level inclusion criteria for the UKB cohort.** Prevalence: proportion of participants with a recorded event on or before the device-wear start date. Incident rate: proportion developing the condition within 6 years among participants without prevalent disease.

| Disease | Prevalent Cases | Prev. N | Prev. (%) | Incident Cases | Inc. N | Inc. Rate (%) |
|---|---|---|---|---|---|---|
| **Cardiovascular** | | | | | | |
| Hypertension | 18,066 | 103,610 | 17.4 | 9,084 | 84,731 | 10.7 |
| Coronary heart disease not otherwise specified | 4,658 | 103,610 | 4.5 | 2,950 | 98,735 | 3.0 |
| Stable angina | 3,277 | 103,610 | 3.2 | 1,523 | 100,204 | 1.5 |
| Atrial fibrillation | 3,034 | 103,610 | 2.9 | 3,137 | 100,364 | 3.1 |
| Myocardial infarction | 2,272 | 103,610 | 2.2 | 968 | 101,261 | 1.0 |
| Venous thromboembolic disease (Excl PE) | 2,114 | 103,610 | 2.0 | 552 | 101,457 | 0.5 |
| Transient ischaemic attack | 1,139 | 103,610 | 1.1 | 729 | 102,409 | 0.7 |
| Pulmonary embolism | 1,022 | 103,610 | 1.0 | 767 | 102,554 | 0.7 |
| Raynaud's syndrome | 992 | 103,610 | 1.0 | 320 | 102,591 | 0.3 |
| Unstable Angina | 992 | 103,610 | 1.0 | 288 | 102,593 | 0.3 |
| Heart failure | 808 | 103,610 | 0.8 | 1,420 | 102,729 | 1.4 |
| Peripheral arterial disease | 736 | 103,610 | 0.7 | 458 | 102,853 | 0.4 |
| Supraventricular tachycardia | 716 | 103,610 | 0.7 | 397 | 102,868 | 0.4 |
| Nonrheumatic mitral valve disorders | 574 | 103,610 | 0.6 | 674 | 102,998 | 0.7 |
| Nonrheumatic aortic valve disorders | 534 | 103,610 | 0.5 | 705 | 103,034 | 0.7 |
| Ischaemic stroke | 426 | 103,610 | 0.4 | 774 | 103,127 | 0.8 |
| Right bundle branch block | 348 | 103,610 | 0.3 | 597 | 103,227 | 0.6 |
| Left bundle branch block | 325 | 103,610 | 0.3 | 513 | 103,257 | 0.5 |
| Multiple valve dz | 271 | 103,610 | 0.3 | 878 | 103,308 | 0.8 |
| Abdominal aortic aneurysm | 238 | 103,610 | 0.2 | 339 | 103,350 | 0.3 |
| Subarachnoid haemorrhage | 224 | 103,610 | 0.2 | 91 | 103,373 | 0.1 |
| Stroke NOS | 211 | 103,610 | 0.2 | 116 | 103,389 | 0.1 |
| Atrioventricular block, first degree | 200 | 103,610 | 0.2 | 485 | 103,383 | 0.5 |
| Other Cardiomyopathy | 180 | 103,610 | 0.2 | 243 | 103,413 | 0.2 |
| Pericardial effusion (noninflammatory) | 179 | 103,610 | 0.2 | 217 | 103,419 | 0.2 |
| Ventricular tachycardia | 162 | 103,610 | 0.2 | 180 | 103,439 | 0.2 |
| Intracerebral haemorrhage | 142 | 103,610 | 0.1 | 165 | 103,459 | 0.2 |
| Dilated cardiomyopathy | 131 | 103,610 | 0.1 | 106 | 103,475 | 0.1 |
| Rheumatic valve dz | 130 | 103,610 | 0.1 | 221 | 103,473 | 0.2 |
| Atrioventricular block, complete | 106 | 103,610 | 0.1 | 220 | 103,496 | 0.2 |
| Sick sinus syndrome | 100 | 103,610 | 0.1 | 139 | 103,503 | 0.1 |
| Atrioventricular block, second degree | 94 | 103,610 | 0.1 | 155 | 103,512 | 0.1 |
| Hypertrophic Cardiomyopathy | 87 | 103,610 | 0.1 | 65 | 103,518 | 0.1 |
| Primary pulmonary hypertension | 82 | 103,610 | 0.1 | 28 | 103,525 | 0.0 |
| Subdural haematoma - nontraumatic | 69 | 103,610 | 0.1 | 70 | 103,540 | 0.1 |
| Secondary pulmonary hypertension | 46 | 103,610 | 0.0 | 190 | 103,557 | 0.2 |
| Trifascicular block | 16 | 103,610 | 0.0 | 40 | 103,593 | 0.0 |
| Bifascicular block | 15 | 103,610 | 0.0 | 51 | 103,593 | 0.0 |
| **Endocrine** | | | | | | |
| Hypo or hyperthyroidism | 7,081 | 103,610 | 6.8 | 1,388 | 96,406 | 1.4 |
| Hypothyroidism | 6,317 | 103,610 | 6.1 | 1,285 | 97,180 | 1.3 |
| Obesity | 4,036 | 103,610 | 3.9 | 3,287 | 99,329 | 3.3 |
| Diabetes Type II | 3,671 | 103,610 | 3.5 | 2,281 | 99,735 | 2.3 |
| Diabetes NOS | 3,480 | 103,610 | 3.4 | 495 | 100,095 | 0.5 |
| Hyperthyroidism | 1,521 | 103,610 | 1.5 | 305 | 102,056 | 0.3 |
| Diabetes Type I | 477 | 103,610 | 0.5 | 178 | 103,121 | 0.2 |
| Thyroiditis unspecified | 303 | 103,610 | 0.3 | 75 | 103,302 | 0.1 |
| Polycystic ovarian syndrome | 292 | 103,610 | 0.3 | 13 | 103,318 | 0.0 |

*Continued on next page*

Supplementary Table 13 – *continued*

| Disease | Prevalent Cases | Prev. N | Prev. (%) | Incident Cases | Inc. N | Inc. Rate (%) |
|---|---|---|---|---|---|---|
| Hyperparathyroidism | 257 | 103,610 | 0.2 | 217 | 103,339 | 0.2 |
| Cystic Fibrosis | 28 | 103,610 | 0.0 | 0 | 103,582 | 0.0 |
| Syndrome of inappropriate secretion of antidiuretic hormone | 20 | 103,610 | 0.0 | 64 | 103,590 | 0.1 |
| **Musculoskeletal** | | | | | | |
| Osteoarthritis (excl spine) | 17,518 | 103,610 | 16.9 | 6,301 | 85,520 | 7.4 |
| Enthesopathies & synovial disorders | 14,486 | 103,610 | 14.0 | 2,184 | 88,752 | 2.5 |
| Carpal tunnel syndrome | 8,238 | 103,610 | 8.0 | 383 | 95,331 | 0.4 |
| Intervertebral disc disorders | 5,126 | 103,610 | 4.9 | 1,338 | 98,381 | 1.4 |
| Spondylosis | 4,081 | 103,610 | 3.9 | 1,564 | 99,403 | 1.6 |
| Fracture of wrist | 3,841 | 103,610 | 3.7 | 861 | 99,697 | 0.9 |
| Osteoporosis | 3,494 | 103,610 | 3.4 | 1,710 | 99,968 | 1.7 |
| Gout | 2,676 | 103,610 | 2.6 | 707 | 100,857 | 0.7 |
| Fibromatoses | 1,614 | 103,610 | 1.6 | 669 | 101,935 | 0.7 |
| Rheumatoid Arthritis | 1,446 | 103,610 | 1.4 | 586 | 102,115 | 0.6 |
| Spinal stenosis | 827 | 103,610 | 0.8 | 953 | 102,717 | 0.9 |
| Polymyalgia Rheumatica | 570 | 103,610 | 0.6 | 526 | 102,997 | 0.5 |
| Spondylolisthesis | 561 | 103,610 | 0.5 | 301 | 103,022 | 0.3 |
| Scoliosis | 537 | 103,610 | 0.5 | 341 | 103,050 | 0.3 |
| Fracture of hip | 495 | 103,610 | 0.5 | 523 | 103,093 | 0.5 |
| Psoriatic arthropathy | 373 | 103,610 | 0.4 | 91 | 103,230 | 0.1 |
| Ankylosing spondylitis | 365 | 103,610 | 0.4 | 47 | 103,239 | 0.0 |
| Collapsed vertebra | 291 | 103,610 | 0.3 | 226 | 103,308 | 0.2 |
| Sjogren's disease | 213 | 103,610 | 0.2 | 104 | 103,387 | 0.1 |
| Lupus erythematosus (local and systemic) | 209 | 103,610 | 0.2 | 43 | 103,397 | 0.0 |
| Giant Cell arteritis | 108 | 103,610 | 0.1 | 128 | 103,492 | 0.1 |
| Systemic sclerosis | 59 | 103,610 | 0.1 | 25 | 103,550 | 0.0 |
| Postinfective and reactive arthropathies | 53 | 103,610 | 0.1 | 13 | 103,557 | 0.0 |
| Juvenile arthritis | 13 | 103,610 | 0.0 | 6 | 103,597 | 0.0 |
| Enteropathic arthropathy | 11 | 103,610 | 0.0 | 3 | 103,599 | 0.0 |
| **Neurological** | | | | | | |
| Migraine | 7,979 | 103,610 | 7.7 | 591 | 95,578 | 0.6 |
| Postviral fatigue syndrome, neurasthenia and fibromyalgia | 2,937 | 103,610 | 2.8 | 185 | 100,655 | 0.2 |
| Peripheral neuropathies (excl. cranial nerve, carpal tunnel syndromes) | 1,993 | 103,610 | 1.9 | 949 | 101,539 | 0.9 |
| Epilepsy | 1,014 | 103,610 | 1.0 | 253 | 102,575 | 0.2 |
| Bell's palsy | 462 | 103,610 | 0.4 | 113 | 103,142 | 0.1 |
| Multiple sclerosis | 408 | 103,610 | 0.4 | 62 | 103,195 | 0.1 |
| Trigeminal neuralgia | 386 | 103,610 | 0.4 | 110 | 103,208 | 0.1 |
| Parkinson's disease | 224 | 103,610 | 0.2 | 295 | 103,374 | 0.3 |
| Essential tremor | 207 | 103,610 | 0.2 | 102 | 103,394 | 0.1 |
| Disorders of autonomic nervous system | 132 | 103,610 | 0.1 | 68 | 103,477 | 0.1 |
| Diabetic neurological complications | 114 | 103,610 | 0.1 | 107 | 103,493 | 0.1 |
| Myasthenia gravis | 48 | 103,610 | 0.0 | 31 | 103,560 | 0.0 |
| Cerebral Palsy | 32 | 103,610 | 0.0 | 5 | 103,578 | 0.0 |
| Motor neuron disease | 30 | 103,610 | 0.0 | 66 | 103,577 | 0.1 |
| Intracranial hypertension | 21 | 103,610 | 0.0 | 7 | 103,589 | 0.0 |
| **Psychiatric** | | | | | | |
| Depression | 48,019 | 103,610 | 46.3 | 531 | 55,565 | 1.0 |
| Anxiety disorders | 26,380 | 103,610 | 25.5 | 1,134 | 77,161 | 1.5 |
| Alcohol Problems | 1,876 | 103,610 | 1.8 | 831 | 101,553 | 0.8 |
| Bipolar affective disorder and mania | 611 | 103,610 | 0.6 | 55 | 102,993 | 0.1 |

*Continued on next page*

Supplementary Table 13 – *continued*

| Disease | Prevalent Cases | Prev. N | Prev. (%) | Incident Cases | Inc. N | Inc. Rate (%) |
|---|---|---|---|---|---|---|
| Obsessive-compulsive disorder | 505 | 103,610 | 0.5 | 11 | 103,104 | 0.0 |
| Anorexia and bulimia nervosa | 201 | 103,610 | 0.2 | 7 | 103,409 | 0.0 |
| Other psychoactive substance misuse | 168 | 103,610 | 0.2 | 56 | 103,434 | 0.1 |
| Schizophrenia, schizotypal and delusional disorders | 165 | 103,610 | 0.2 | 58 | 103,440 | 0.1 |
| Personality disorders | 116 | 103,610 | 0.1 | 20 | 103,493 | 0.0 |
| Autism and Asperger's syndrome | 91 | 103,610 | 0.1 | 12 | 103,518 | 0.0 |
| Dementia | 52 | 103,610 | 0.1 | 466 | 103,552 | 0.5 |
| Alzheimer's disease | 33 | 103,610 | 0.0 | 197 | 103,572 | 0.2 |
| Delirium, not induced by alcohol and other psychoactive substances | 33 | 103,610 | 0.0 | 597 | 103,564 | 0.6 |
| Dementia (excluding Alzheimer's) | 24 | 103,610 | 0.0 | 343 | 103,583 | 0.3 |
| Intellectual disability | 11 | 103,610 | 0.0 | 11 | 103,598 | 0.0 |
| Hyperkinetic disorders | 2 | 103,610 | 0.0 | 6 | 103,607 | 0.0 |
| **Respiratory** | | | | | | |
| Allergic and chronic rhinitis | 30,298 | 103,610 | 29.2 | 632 | 73,206 | 0.9 |
| Asthma | 14,959 | 103,610 | 14.4 | 983 | 88,558 | 1.1 |
| Chronic sinusitis | 7,052 | 103,610 | 6.8 | 456 | 96,471 | 0.5 |
| COPD | 2,126 | 103,610 | 2.1 | 1,363 | 101,391 | 1.3 |
| COPD_excl_bronchitis_NOS | 2,047 | 103,610 | 2.0 | 1,323 | 101,478 | 1.3 |
| Nasal polyp | 1,740 | 103,610 | 1.7 | 174 | 101,847 | 0.2 |
| Sleep apnoea | 1,341 | 103,610 | 1.3 | 683 | 102,213 | 0.7 |
| Bronchiectasis | 806 | 103,610 | 0.8 | 487 | 102,770 | 0.5 |
| Pleural effusion | 674 | 103,610 | 0.7 | 1,512 | 102,873 | 1.5 |
| Pneumothorax | 479 | 103,610 | 0.5 | 200 | 103,123 | 0.2 |
| Hypertrophy of nasal turbinates | 433 | 103,610 | 0.4 | 71 | 103,162 | 0.1 |
| Pulmonary collapse (excl pneumothorax) | 278 | 103,610 | 0.3 | 933 | 103,291 | 0.9 |
| Other interstitial pulmonary diseases with fibrosis | 189 | 103,610 | 0.2 | 289 | 103,412 | 0.3 |
| Respiratory failure | 172 | 103,610 | 0.2 | 727 | 103,412 | 0.7 |
| Pleural plaque | 106 | 103,610 | 0.1 | 193 | 103,495 | 0.2 |
| Asbestosis | 47 | 103,610 | 0.0 | 37 | 103,562 | 0.0 |
| Aspiration pneumonitis | 39 | 103,610 | 0.0 | 257 | 103,557 | 0.2 |

Supplementary Table 14: **Supervised fine-tuning model performance.** Participant-level logistic-probe AUROC results on the fixed pilot validation set. Initial convolutional blocks were frozen for all models. Only layers downstream of the feature extractor were fine-tuned.

| | | | Pretrained initialisation | | Random init. after Conv |
|---|---|---|---|---|---|
| Organ system | Disease | Cases, *n* | Disease-specific | Shared six-organ | Disease-specific |
| Cardiovascular | Hypertension | 2,580 | 0.742 | 0.743 | 0.730 |
| Endocrine | Type 2 diabetes | 510 | 0.822 | 0.829 | 0.793 |
| Musculoskeletal | Osteoarthritis | 2,439 | 0.698 | 0.698 | 0.685 |
| Neurological | Migraine | 1,118 | 0.632 | 0.638 | 0.625 |
| Psychiatric | Depression | 6,966 | 0.627 | 0.630 | 0.608 |
| Respiratory | Asthma | 2,152 | 0.622 | 0.615 | 0.598 |
| Average | – | – | 0.691 | 0.692 | 0.673 |

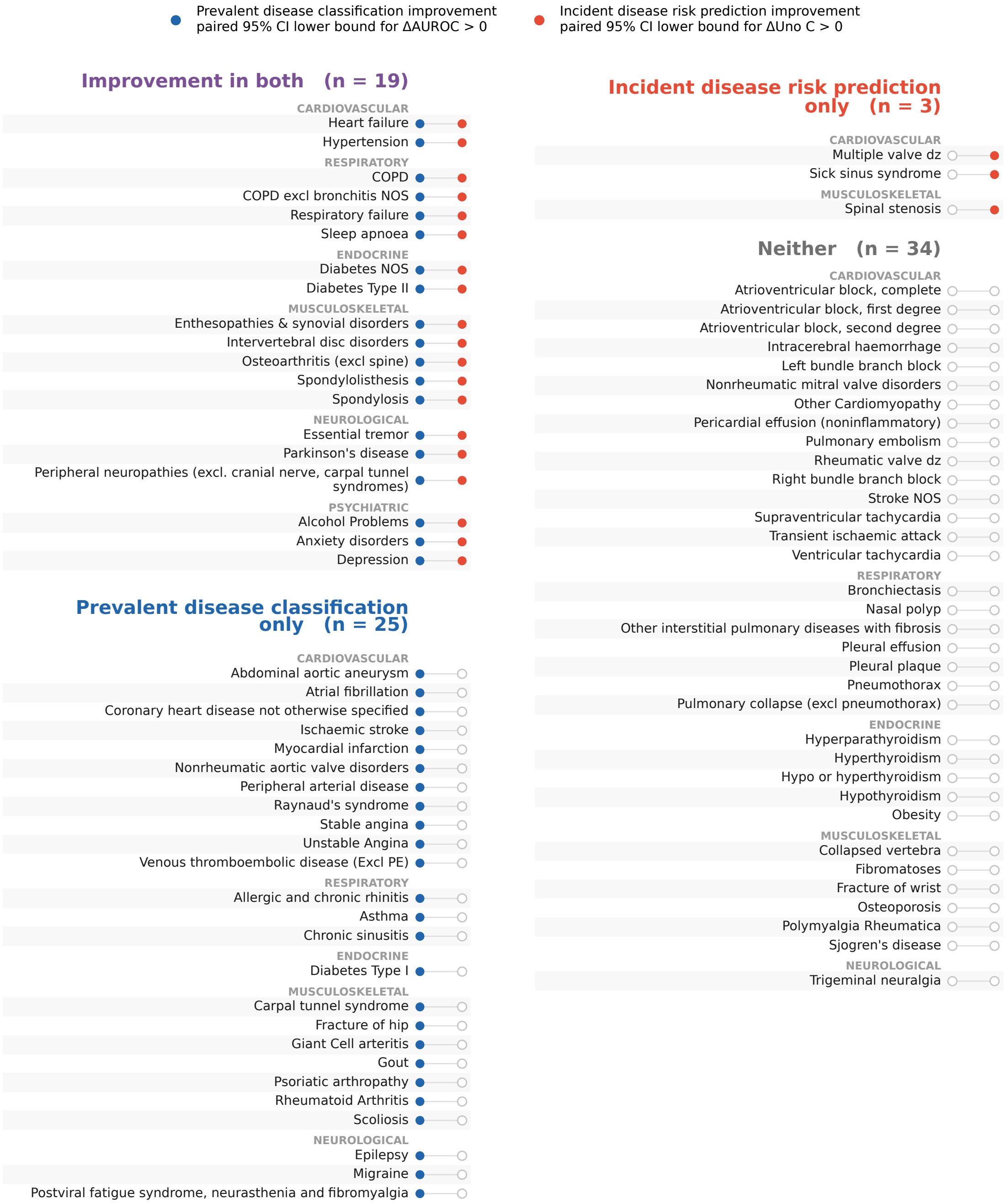


Supplementary Figure 8: **Overlap in positive performance differences across disease prediction tasks.** Overlap between diseases for which adding Sensori embeddings to the clinical covariate model yielded a positive performance difference in prevalent disease classification (delta AUROC > 0) and six-year incident disease risk prediction (delta Uno's C-index > 0).

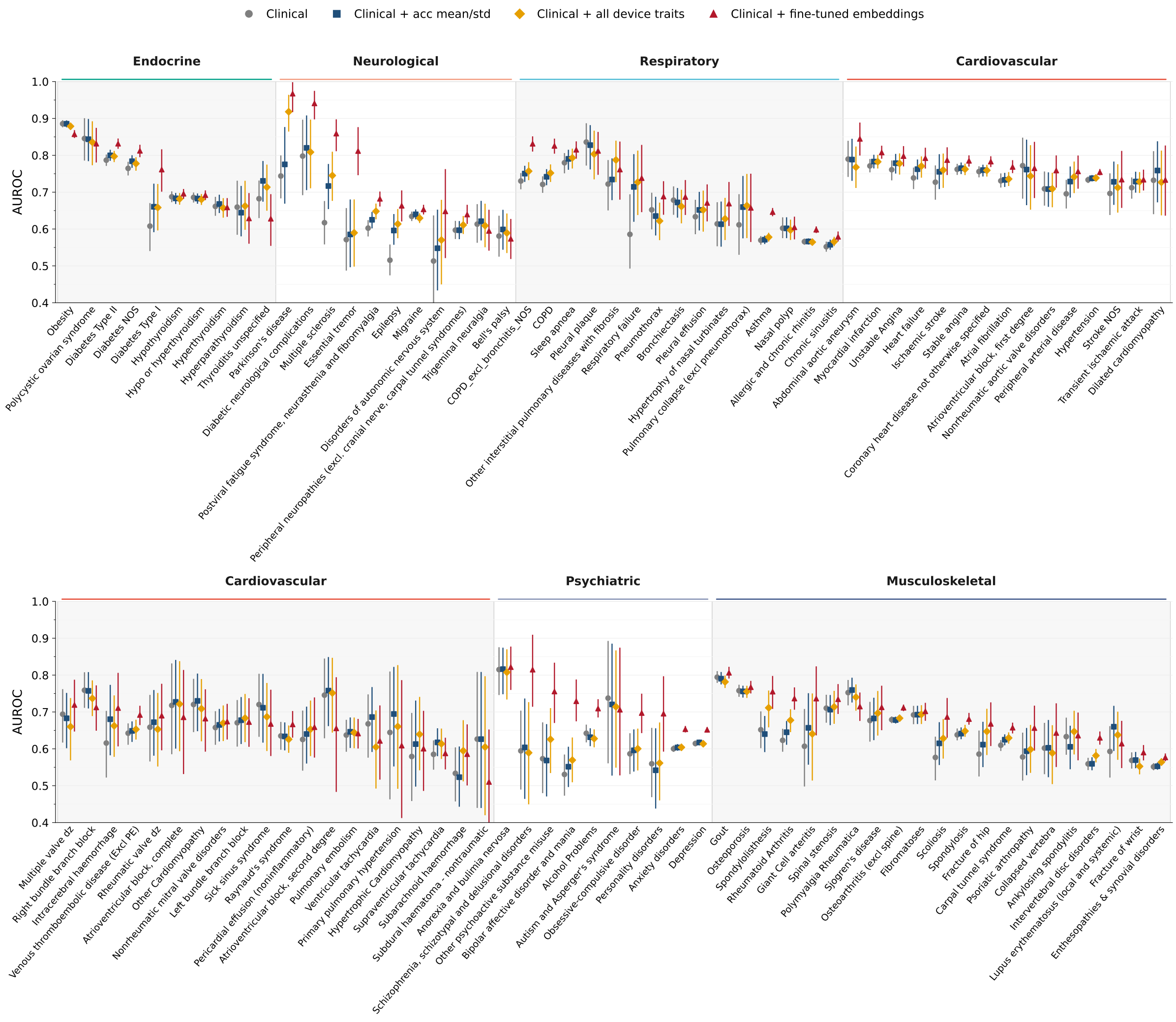


Supplementary Figure 9: **Comparison of prevalent disease classification performance.** Performance was compared across four models using AUROC: the clinical covariate model alone, the clinical covariate model with acceleration statistics, the clinical covariate model with conventional device-measured behavioural traits and the clinical covariate model with Sensori embeddings. Error bars show 95% CIs in the UKB held-out test set.

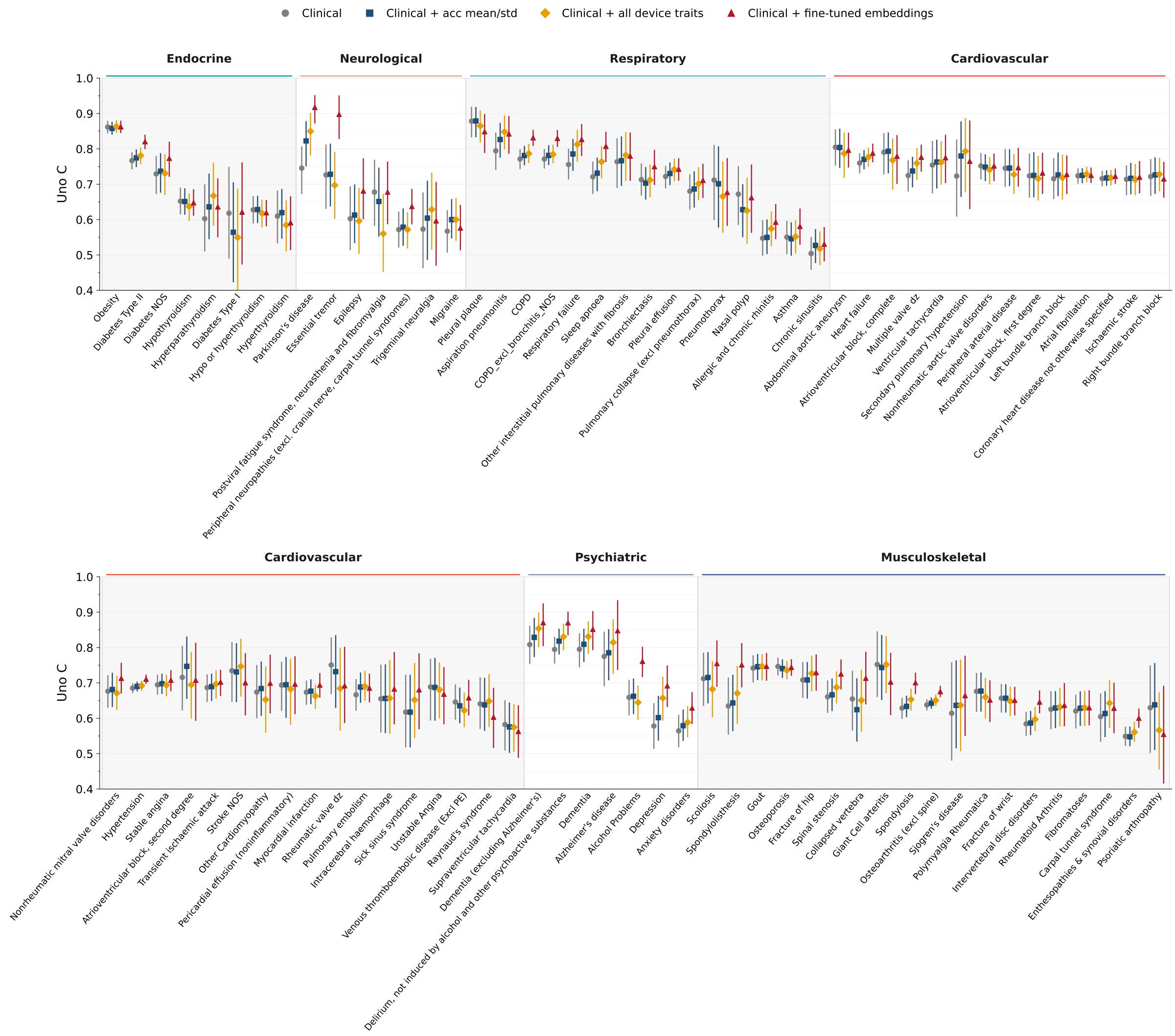


Supplementary Figure 10: **Comparison of incident disease risk prediction performance.** Performance was compared across four models using Uno's C-index: the clinical covariate model alone, the clinical covariate model with acceleration statistics, the clinical covariate model with conventional device-measured behavioural traits and the clinical covariate model with Sensori embeddings. Error bars show 95% CIs in the UKB held-out test set.

Supplementary Table 15: **Paired-bootstrap comparison of clinical covariates plus fine-tuned Sensori embeddings against alternative feature models for disease prediction.**

| Comparator | delta mean (median) | Improved *n* (%) | Worsened *n* (%) | Insignificant *n* (%) |
|---|---|---|---|---|
| **a, Prevalent disease classification (delta AUROC)** | | | | |
| Clinical | 0.044 (0.036) | 52 (51.0) | 3 (2.9) | 47 (46.1) |
| Clinical + Acc mean/s.d. | 0.031 (0.026) | 47 (46.1) | 4 (3.9) | 51 (50.0) |
| Clinical + 18 device-measured behavioural traits | 0.028 (0.022) | 47 (46.1) | 2 (2.0) | 53 (52.0) |
| **b, Incident disease risk prediction (delta Uno's C-index)** | | | | |
| Clinical | 0.027 (0.021) | 26 (29.9) | 0 (0.0) | 61 (70.1) |
| Clinical + Acc mean/s.d. | 0.021 (0.015) | 23 (26.4) | 0 (0.0) | 64 (73.6) |
| Clinical + 18 device-measured behavioural traits | 0.021 (0.014) | 22 (25.3) | 0 (0.0) | 65 (74.7) |



\end{document}